\PassOptionsToPackage{dvipsnames}{xcolor}
\documentclass{article} 
\usepackage{iclr2025_conference,times}

\usepackage{amsmath,amsfonts,bm}

\def\eqref#1{equation~\ref{#1}}

\def\1{\bm{1}}

\DeclareMathAlphabet{\mathsfit}{\encodingdefault}{\sfdefault}{m}{sl}
\SetMathAlphabet{\mathsfit}{bold}{\encodingdefault}{\sfdefault}{bx}{n}

\usepackage[utf8]{inputenc} 
\usepackage[T1]{fontenc}    
\usepackage{hyperref}       
\usepackage{url}            
\usepackage{booktabs}       
\usepackage{amsfonts}       
\usepackage{nicefrac}       
\usepackage{microtype}      
\usepackage{colortbl}
\usepackage{makecell}

\usepackage{amsmath}
\usepackage{cleveref}
\usepackage{pifont}
\usepackage{enumitem}
\usepackage{multicol}
\usepackage[most]{tcolorbox}
\usepackage{multirow}

\usepackage{amsthm}
\usepackage{amssymb}
\newtheorem{definition}{Definition}
\usepackage[toc,page]{appendix}

\usepackage{wrapfig}

\usepackage{xspace}

\newcommand{\ourmethod}{MoG\xspace}

\hypersetup{
    colorlinks=true,
    linkcolor=red,
    citecolor=cyan,
    filecolor=magenta,      
    urlcolor=magenta,
    }

\usepackage[ruled]{algorithm2e}
\SetKwInOut{Input}{Input}\SetKwInOut{Output}{Output}
\SetKwComment{Comment}{$\triangleright$\ }{}

\title{Graph Sparsification via Mixture of Graphs}

\author{Guibin Zhang$^{1\dagger}$,\; 
Xiangguo Sun$^{2\dagger}$,\; 
Yanwei Yue$^{1\dagger}$,\; 
Chonghe Jiang$^{2}$,\\
\textbf{Kun Wang}$^{3}$,\;
\textbf{Tianlong Chen}$^{4}$,
\textbf{Shirui Pan}$^{5}$
	\\
	$^{1}$Tongji University\;
    $^{2}$CUHK \;
    $^{3}$NTU\;
    $^{4}$UNC - Chapel Hill\;
    $^{5}$Griffith University \;
}

\iclrfinalcopy 
\begin{document}

\maketitle

\begin{abstract}
\vspace{-0.5em}
Graph Neural Networks (GNNs) have demonstrated superior performance across various graph learning tasks but face significant computational challenges when applied to large-scale graphs. One effective approach to mitigate these challenges is graph sparsification, which involves removing non-essential edges to reduce computational overhead. However, previous graph sparsification methods often rely on a single global sparsity setting and uniform pruning criteria, failing to provide customized sparsification schemes for each node's complex local context.
In this paper, we introduce Mixture-of-Graphs (MoG), leveraging the concept of Mixture-of-Experts (MoE), to dynamically select tailored pruning solutions for each node. Specifically, MoG incorporates multiple sparsifier experts, each characterized by unique sparsity levels and pruning criteria, and selects the appropriate experts for each node. Subsequently, MoG performs a mixture of the sparse graphs produced by different experts on the Grassmann manifold to derive an optimal sparse graph. One notable property of MoG is its entirely local nature, as it depends on the specific circumstances of each individual node. Extensive experiments on four large-scale OGB datasets and two superpixel datasets, equipped with five GNN backbones, demonstrate that MoG (I) identifies subgraphs at higher sparsity levels ($8.67\%\sim 50.85\%$), with performance equal to or better than the dense graph, (II) achieves $1.47-2.62\times$ speedup in GNN inference with negligible performance drop, and (III) boosts ``top-student'' GNN performance ($1.02\%\uparrow$ on RevGNN+\textsc{ogbn-proteins} and $1.74\%\uparrow$ on DeeperGCN+\textsc{ogbg-ppa}).  The source code is 
anonymously available at \url{https://github.com/yanweiyue/MoG}.
\end{abstract}

\section{Introduction}
\vspace{-0.6em}
Graph Neural Networks (GNNs) \citep{sun2023all,zhou2020graph} have become prominent for confronting graph-related learning tasks, including social recommendation~\citep{wu2021self,yu2022graph}, fraud detection~\citep{sun2022structure, wang2019semi,cheng2020graph}, drug design~\citep{zhang2023learning}, and many others~\citep{wu2023graph, sun2023self}. The superiority of GNNs stems from iterative \emph{aggregation} and \emph{update} processes. The former accumulates embeddings from neighboring nodes via sparse matrix-based operations (\textit{e.g.}, sparse-dense matrix multiplication ({\fontfamily{lmtt}\selectfont \textbf{SpMM}}) and sampled dense-dense matrix multiplication ({\fontfamily{lmtt}\selectfont \textbf{SDDMM}})~\citep{fey2019fast,wang2019deep}), and the latter updates the central nodes' embeddings using dense matrix-based operations (\textit{e.g.}, {\fontfamily{lmtt}\selectfont \textbf{MatMul}}) \citep{fey2019fast,wang2019deep}. {\fontfamily{lmtt}\selectfont \textbf{SpMM}} typically contributes the most substantial part ($\sim$70\%) to the computational demands \citep{liu2023dspar,zhang2024two}, influenced largely by the graph's scale. Nevertheless, large-scale graphs are widespread in real-world scenarios \citep{wang2022searching,jin2021graph,zhang2024graph}, leading to substantial computational burdens, which hinder the efficient processing of features during the training and inference, posing headache barriers to deploying GNNs in the limited resources environments.

\begin{figure*}
\centering
\includegraphics[width=\linewidth]{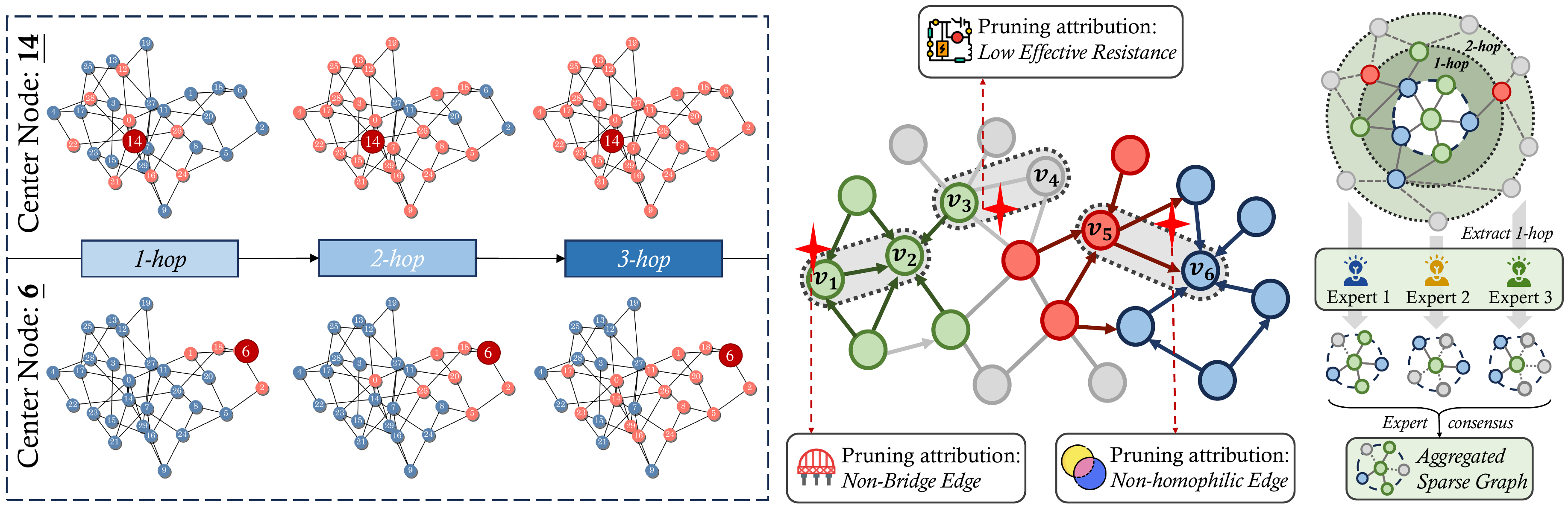}
\vspace{-6mm}
\caption{(\textbf{\textit{Left}}) We illustrated the $k$-hop neighborhood expansion rates for nodes 6 and 14, which is proportional to the amount of message they receive as the GNN layers deepen; (\textbf{\textit{Middle}}) The local patterns of different nodes vary, hence the attributions of edge pruning may also differ. For instance, pruning $(v_1,v_2)$ might be due to its non-bridge identity, while pruning $(v_5,v_6)$ could be attributed to it non-homophilic nature; (\textbf{\textit{Right}}) The overview of our proposed MoG.}
\label{fig:intro}
\vspace{-1em}
\end{figure*}

To conquer the above challenge, graph sparsification~\citep{chen2023demystifying,hashemi2024comprehensive} has recently seen a revival as it directly reduces the \textit{aggregation} process associated with {\fontfamily{lmtt}\selectfont \textbf{SpMM}}~\citep{liu2023dspar,zhang2024two} in GNNs. Specifically, graph sparsification is a technique that approximates a given graph by creating a sparse subgraph with a subset of vertices and/or edges. Since the execution time of {\fontfamily{lmtt}\selectfont \textbf{SpMM}} is directly related to the number of edges in the graph, this method can significantly accelerate GNN training or inference. Existing efforts such as UGS~\citep{chen2021unified}, DSpar~\citep{liu2023dspar}, and AdaGLT~\citep{zhang2023graph} have achieved notable successes, with some maintaining GNN performance even with up to 40\% edge sparsity.

Beyond serving as a \textbf{computational accelerator}, the purpose of graph sparsification extends further. Another research line leverages graph sparsification as a \textbf{performance booster} to remove task-irrelevant edges and pursue highly performant and robust GNNs~\citep{zheng2020robust}. Specifically, it is argued that due to uncertainty and complexity in data collection, graph structures are inevitably redundant, biased, and noisy~\citep{li2024gslb}. Therefore, employing graph sparsification can effectively facilitate the evolution of graph structures towards cleaner conditions~\citep{zheng2020robust,luo2021learning}, and finally boost GNN performance.

However, existing sparsification methods, namely \textit{sparsifiers}, whether aimed at achieving higher sparsity or seeking enhanced performance, often adopt a rigid, global approach to conduct graph sparsification, thus suffering from the inflexibility in two aspects:

\vspace{-0.7em}
\begin{itemize}[leftmargin=*]
    \item[\ding{182}] \textbf{\textit{Inflexibility of sparsity level.}} Previous sparsifiers globally score all edges uniformly and prune them based on a preset sparsity level~\citep{chen2023demystifying}. However, as shown in \Cref{fig:intro} (\textit{Left}), the degrees of different nodes vary, which leads to varying rates of $k$-hop neighborhood expansion. This phenomenon, along with prior work on node-wise aggregation~\citep{lai2020policy,wang2023snowflake}, suggests that \textit{different nodes require customized sparsity levels tailored to their specific connectivity and local patterns}.

    \item[\ding{183}] \textbf{\textit{Inflexibility of sparsity criteria.}} Previous sparsifiers often operate under a unified guiding principle, such as pruning non-bridge edges \citep{wang2022pruning}, non-homophilic edges \citep{gong2023beyond}, or edges with low effective resistance \citep{spielman2008graph,liu2023dspar}, among others. However, as illustrated in \Cref{fig:intro} (\textit{Middle}), the context of different nodes varies significantly, leading to varied rationales for edge pruning. Therefore, it is essential to  \textit{select appropriate pruning criteria tailored to the specific circumstances of each node to customize the pruning process effectively.}
\end{itemize}
\vspace{-0.7em}

Based on these observations and reflections, we propose the following challenge: \textit{Can we customize the sparsity level and pruning criteria for each node, in the meanwhile ensuring the efficiency of graph sparsification?} Towards this end, we propose a novel graph sparsifier dubbed \underline{M}ixture \underline{o}f \underline{G}raphs (\textbf{MoG}). It comprises multiple \textit{sparsifier experts}, each equipped with distinct pruning criteria and sparsity settings, as in \Cref{fig:intro} (\textit{Right}). Throughout the training process, \ourmethod dynamically selects the most suitable sparsifier expert for each node based on its neighborhood properties. This fosters specialization within each MoG expert, focusing on specific subsets of nodes with similar neighborhood contexts.
After each selected expert prunes the 1-hop subgraph of the central nodes and outputs its sparse version, MoG seamlessly integrates these sparse subgraphs on the Grassmann manifold in an expert-weighted manner, thereby forming an optimized sparse graph.

We validate the effectiveness of MoG through a comprehensive series of large-scale tasks. Experiments conducted across six datasets and three GNN backbones showcase that MoG can \ding{182} effectively locate well-performing sparse graphs, maintaining GNN performance losslessly at satisfactory graph sparsity levels ($8.67\%\sim50.85\%$), and even only experiencing a $1.65\%$ accuracy drop at $69.13\%$ sparsity on \textsc{ogbn-proteins}; \ding{183} achieve a tangible $1.47\sim 2.62\times$ inference speedup with negligible performance drop; and \ding{184} boost ROC-AUC by $1.81\%$ on \textsc{ogbg-molhiv}, $1.02\%$ on ogbn-proteins and enhances accuracy by $0.95\%$ on \textsc{ogbn-arxiv} compared to the vanilla backbones.

\section{Technical Backgound}
\vspace{-0.6em}
\paragraph{Notations \& Problem Formulation}\label{para:notation} We consider an undirected graph $\mathcal{G}=\{\mathcal{V},\mathcal{E}\}$, with $\mathcal{V}$ as the node set and $\mathcal{E}$ the edge set. The node features of $\mathcal{G}$ is represented as $\mathbf{X} \in \mathbb{R}^{N \times F}$, where $N = |\mathcal{V}|$ signifies the total number of nodes in the graph. The feature vector for each node $v_i \in \mathcal{V}$, with $F$ dimensions, is denoted by $x_i = \mathbf{X}[i,\cdot]$.  An adjacency matrix $\mathbf{A} \in \{0,1\}^{N \times N}$ is utilized to depict the inter-node connectivity, where $\mathbf{A}[i,j] = 1$ indicates an edge $e_{ij} \in \mathcal{E}$, and $0$ otherwise. For our task of graph sparsification, the core objective is to identify a subgraph $\mathcal{G}_{\text{sub}}$ given a sparsity ratio $s\%$:
\begin{equation}
\mathcal{G}^{\text{sub}} = \{\mathcal{V}, \mathcal{E} \setminus \mathcal{E}'\},\; s\% = \frac{|\mathcal{E}'|}{ |\mathcal{E}|},
\end{equation}
where $\mathcal{G}^{\text{sub}}$ only modifies the edge set $\mathcal{E}$ without altering the node set $\mathcal{V}$, and $\mathcal{E}'$ denotes the removed edges, and $s\%$ represents the ratio of removed edges.
\vspace{-0.5em}
\paragraph{Graph Neural Networks} Graph neural networks (GNNs)~\citep{wu2020comprehensive} have become pivotal for learning graph representations, achieving benchmark performances in various graph tasks at {node-level}~\citep{xiao2022graph}, {edge-level}~\citep{sun2021multi}, and {graph-level}~\citep{liu2022graph}. At the {node-level}, two of the most famous frameworks are GCN~\citep{kipf2016semi} and GraphSAGE~\citep{hamilton2017inductive}, which leverages the message-passing neural network (MPNN) framework~\citep{gilmer2017neural} to aggregate and update node information iteratively. For {edge-level} and {graph-level} tasks, GCN and GraphSAGE can be adapted by simply incorporating a predictor head or pooling layers. Nevertheless, there are still specialized frameworks like SEAL~\citep{zhang2018link} and Neo-GNN~\citep{yun2021neo} for link prediction, and DiffPool~\citep{ying2018hierarchical} and PNA~\citep{corso2020principal} for graph classification. Regardless of the task, MPNN-style GNNs generally adhere to the following paradigm:
\vspace{-0.1em}
\begin{equation}\label{eq:gnn}
\mathbf{h}_i^{(l)} = \text{\fontfamily{lmtt}\selectfont \textbf{COMB}}\left( \mathbf{h}_i^{(l-1)}, \text{\fontfamily{lmtt}\selectfont \textbf{AGGR}}\{  \mathbf{h}_j^{(k-1)}: v_j \in \mathcal{N}(v_i) \} \right),\;0\leq l \leq L
\end{equation}
where $L$ is the number of GNN layers, $\mathbf{h}_i^{(0)} = \mathbf{x}_i$, and $\mathbf{h}_i^{(l)} (1\leq l\leq L)$ denotes $v_i$'s node embedding at the $l$-th layer. {\fontfamily{lmtt}\selectfont \textbf{AGGR}}($\cdot$) and {\fontfamily{lmtt}\selectfont \textbf{COMB}}($\cdot$) represent functions used for aggregating neighborhood information and combining ego- and neighbor-representations, respectively. 
\vspace{-0.5em}
\paragraph{Graph Sparsification} 
 Graph sparsification methods can be categorized by their utility into two main types: {computational accelerators} and {performance boosters}. Regarding {computational accelerators}, early works aimed at speeding up traditional tasks like graph partitioning/clustering often provide theoretical assurances for specific graph properties, such as pairwise distances~\citep{althofer1990generating}, cuts~\citep{abboud2022friendly}, eigenvalue distribution~\citep{batson2013spectral}, and effective resistance~\citep{spielman2008graph}. More contemporary efforts focus on the GNN training and/or inference acceleration, including methods like SGCN~\citep{li2020sgcn}, GEBT~\citep{you2022early}, UGS~\citep{chen2021unified}, DSpar~\citep{liu2023dspar}, and AdaGLT~\citep{zhang2024graph}. Regarding {performance boosters}, methods like NeuralSparse~\citep{zheng2020robust} and PTDNet~\citep{luo2021learning} utilize parameterized denoising networks to eliminate task-irrelevant edges. SUBLIME~\citep{liu2022towards} and Nodeformer~\citep{wu2022nodeformer} also involve refining or inferring a cleaner graph structure followed by $k$-nearest neighbors ($k$NN) sparsification.
\vspace{-0.5em}
\paragraph{Mixture of Experts} The Mixture of Experts (MoE) concept~\citep{jacobs1991adaptive} traces its origins to several seminal works~\citep{chen1999improved,jordan1994hierarchical}. Recently, the sparse MoE architecture~\citep{shazeer2017outrageously,lepikhin2020gshard,fedus2021switch,clark2022unified} has regained attention due to its capacity to support the creation of vast (language) models with trillions of parameters~\citep{clark2022unified,hoffmann2022training}. Given its stability and generalizability, sparse MoE is now broadly implemented in modern frameworks across various domains, including vision~\citep{riquelme2021scaling}, multi-modal~\citep{mustafa2022multimodal}, and multi-task learning~\citep{ma2018modeling,zhu2022uni}. As for graph learning, MoE has been explored for applications in graph classification~\citep{hu2022graphdive}, scene graph generation~\citep{zhou2022came}, molecular representation~\citep{kim2023learning}, graph fairness~\citep{liu2023fair}, and graph diversity modeling~\citep{wang2024graph}.

\section{Methodology}
\subsection{Overview}
\begin{figure}[!t]
\setlength{\abovecaptionskip}{6pt}
\centering
\includegraphics[width=\textwidth]{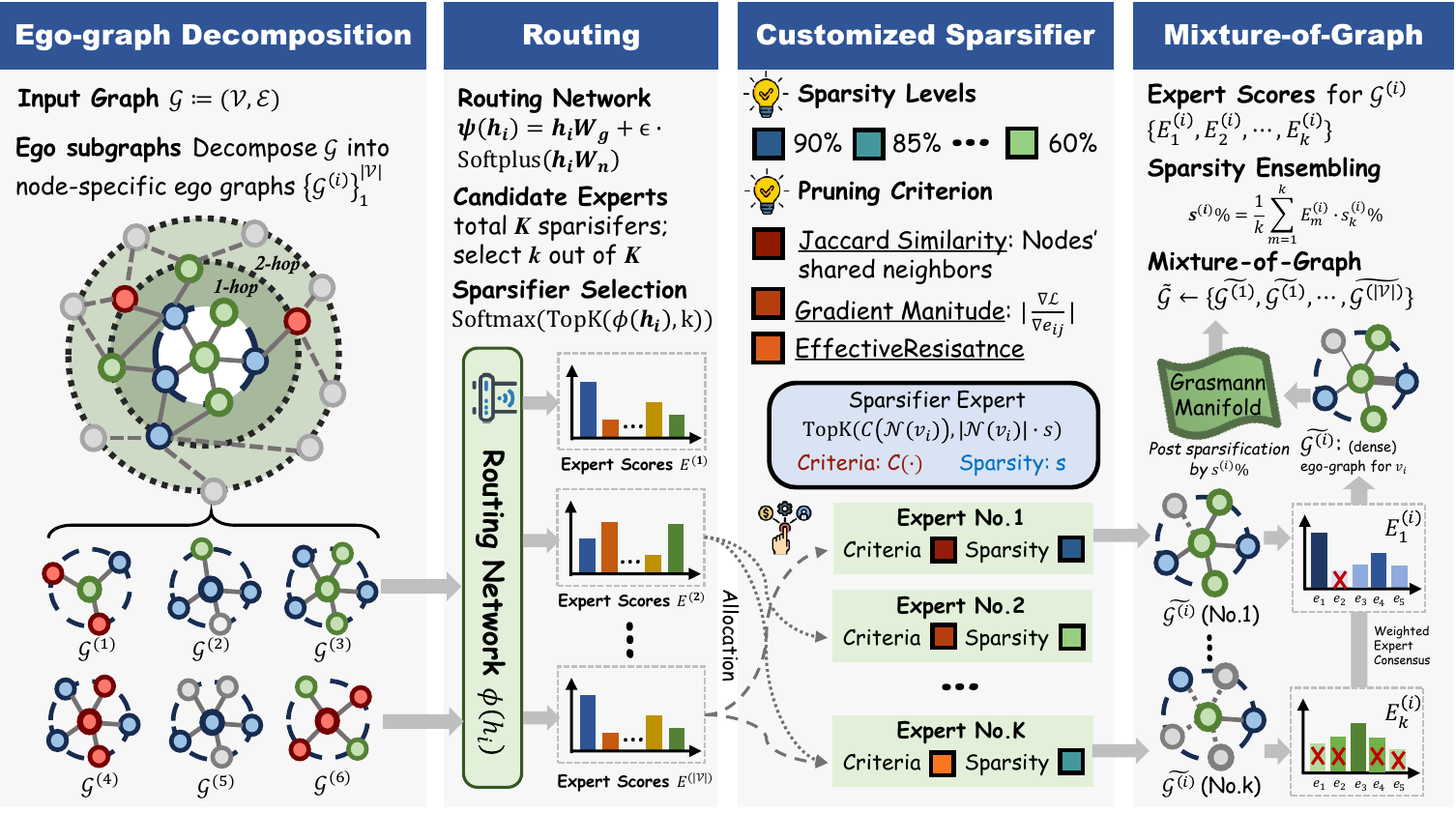}
\vspace{-1.2em}
\caption{The overview of our proposed method. \ourmethod primarily comprises ego-graph decomposition, expert routing, sparsifier customization, and the final graph mixture. For simplicity, we only showcase three pruning criteria including Jaccard similarity, gradient magnitude, and effective resistance.} \label{fig:framework}
\vspace{-0.9em}
\end{figure}

\Cref{fig:framework} illustrates the workflow of our proposed \ourmethod. Specifically, for an input graph, MoG first decomposes it into 1-hop ego graphs for each node. For each node and its corresponding ego graph, a routing network calculates the expert scores. Based on the router's decisions, sparsifier experts with different sparsity levels and pruning criteria are allocated to different nodes. Ultimately, a mixture of graphs is obtained based on the weighted consensus of the sparsifier experts. In the following sections, we will first detail how to route different sparsifiers in \Cref{sec:routing}, then describe how to explicitly model various sparsifier experts in \Cref{sec:sparsifier} and how to ensemble the sparse graphs output by sparsifiers on the Grassmann manifold in \Cref{subsec:gmgm}. Finally, the overall optimization process and complexity analysis of \ourmethod is placed in \Cref{sec:optim}.

\vspace{-0.5em}
\subsection{Routing to Diverse Experts}\label{sec:routing}
Following the classic concept of a (sparsely-gated) mixture-of-experts~\citep{zhao2024sparse}, which assigns the most suitable expert(s) to each input sample, \ourmethod aims to allocate the most appropriate sparsity level and pruning criteria to each input node. To achieve this, we first decompose the input graph $\mathcal{G}=\{\mathcal{V},\mathcal{E}\}$ into 1-hop ego graphs centered on different nodes, denoted as $\{\mathcal{G}^{(1)}, \mathcal{G}^{(2)}, \cdots, \mathcal{G}^{(N)}\}$, where $\mathcal{G}^{(i)}=\{\mathcal{V}^{(i)},\mathcal{E}^{(i)}\}$, $\mathcal{V}^{(i)}=\{v_j|v_j\in\mathcal{N}(v_i)\}$, $\mathcal{E}^{(i)}=\{e_{ij}|(v_i,v_j)\in\mathcal{E}\}$. Assuming we have $K$ sparsifier experts, for each node $v_i$ and its corresponding ego graph $\mathcal{G}^{(i)}$, we aim to select $k$ most suitable sparsifiers. We employ the noisy top-$k$ gating mechanism following~\citet{shazeer2017outrageously}:
\begin{align}\label{eq:routing}
    \Psi(\mathcal{G}^{(i)}) = \operatorname{Softmax}(\operatorname{TopK}(\psi(x_i),k)),\\
    \psi(x_i) = x_iW_g + \epsilon\cdot\operatorname{Softplus}(x_iW_n),
\end{align}
where $\psi(x_i) \in \mathbb{R}^K$ is the calculated scores of $v_i$ for total $K$ experts, $\operatorname{TopK}(\cdot)$ is a selection function that outputs the largest $k$
values, and $\Psi(\mathcal{G}^{(i)}) \in \mathbb{R}^k = [E^{(i)}_1, E^{(i)}_2, \cdots,E^{(i)}_k]$ represents those for selected $k$ experts. In $\Psi(\mathcal{G}^{(i)})$, $\epsilon\in\mathcal{N}(0,1)$ denotes the standard Gaussian noise, $W_g\in\mathbb{R}^{K\times F}$ and $W_n\in\mathbb{R}^{K\times F}$ are trainable parameters that learn clean and noisy scores, respectively. 

After determining the appropriate experts, we proceed to generate different sparse graphs with diverse sparsifiers. We denote each sparsifier by $\kappa(\cdot)$, which takes in a dense graph $\mathcal{G}$ and outputs a sparse one $\Tilde{\mathcal{G}}=\kappa(\mathcal{G})$. Based on this, for each node $v_i$ and its ego graph $\mathcal{G}^{(i)}$, the routing network selects $k$ experts that produce $k$ sparse ego graphs. Notably, sparsifiers differ in their pruning rates (\textit{i.e.} the proportion of the edges to be removed) and the pruning criteria, which will be detailed in \Cref{sec:sparsifier}. \ourmethod's dynamic selection of different sparsifiers for each node aids in identifying pruning strategies truly adapted to the node's local context. Formally, the mixture of $k$ sparse graphs can be written as:
\begin{equation}\label{eq:esmb_1}
\widehat{\mathcal{G}^{(i)}} = \text{\fontfamily{lmtt}\selectfont \textbf{ESMB}}(\{\widetilde{\mathcal{G}}^{(i)}_{m}\}_{m=1}^{k}),\;\; \widetilde{\mathcal{G}}^{(i)}_{m} = \kappa^m(\mathcal{G}^{(i)}),
\end{equation}
where $\text{\fontfamily{lmtt}\selectfont \textbf{ESMB}}(\cdot)$ is a combination function that receives $k$ sparse graphs and ideally outputs an emsemble version $\widehat{\mathcal{G}^{(i)}}=\{\widehat{\mathcal{V}^{(i)}},\widehat{\mathcal{E}^{(i)}}\}$ that preserves their desirable properties. 
It is noteworthy that, \ourmethod can \textit{seamlessly} integrate with any GNN backbone after obtaining each node's sparse ego graph. Specifically, we modify the aggregation method in \Cref{eq:gnn} as follows:
\begin{equation}\label{eq:gnn2}
\mathbf{h}_i^{(l)} = \text{\fontfamily{lmtt}\selectfont \textbf{COMB}}\left( \mathbf{h}_i^{(l-1)}, \text{\fontfamily{lmtt}\selectfont \textbf{AGGR}}\{  \mathbf{h}_j^{(k-1)}: v_j \in \widehat{\mathcal{V}^{(i)}} \} \right).
\end{equation}
\ourmethod acts as a plug-and-play module that can be pre-attached to any GNN architecture, leveraging multi-expert sparsification to enhance GNNs with (1) performance improvements from removing task-irrelevant edges (validated in \Cref{sec:exp_boost}); (2) resistance to high graph sparsity through precise and customized sparsification (validated in \Cref{sec:exp_spar}). The remaining questions now are: \textit{how can we design explicitly different sparsifiers?} and further, \textit{how can we develop an effective combination function that integrates the sparse graphs from different experts?}

\vspace{-0.5em}
\subsection{Customized Sparsifier Modeling}\label{sec:sparsifier}
With the workflow of \ourmethod in mind, in this section, we will delve into how to design sparsifiers driven by various pruning criteria and different levels of sparsity. Revisiting graph-related learning tasks, their objective can generally be considered as learning \( P(\mathbf{Y}|\mathcal{G}) \), which means learning the distribution of the target \( \mathbf{Y} \) given an input graph. Based on this, a sparsifier \( \kappa(\cdot) \) can be formally expressed as follows:
\begin{equation}\label{eq:aim}
\begin{aligned}
P(\mathbf{Y}|\mathcal{G})
& \approx \sum_{g\in\mathbb{S}_{\mathcal{G}}}P(\mathbf{Y}\;|\;\Tilde{\mathcal{G}})P(\Tilde{\mathcal{G}}\;|\;\mathcal{G})
& \approx  \sum_{g\in\mathbb{S}_{\mathcal{G}}} Q_\Theta(\mathbf{Y}\;|\;\Tilde{\mathcal{G}})Q_\kappa(\Tilde{\mathcal{G}}\;|\;\mathcal{G})
\end{aligned}
\end{equation}
where $\mathbb{S}_\mathcal{G}$ is a class of sparsified subgraphs of $\mathcal{G}$. The second term in \Cref{eq:aim} aims to approximate the distribution of \( \mathbf{Y} \) using the sparsified graph \( \Tilde{\mathcal{G}} \) as a bottleneck, while the third term uses two approximation functions \( Q_\Theta \) and \( Q_\kappa \) for \( P(\mathbf{Y} \;|\; \Tilde{\mathcal{G}}) \) and \( P(\Tilde{\mathcal{G}} \;|\; \mathcal{G}) \) parameterized by $\Theta$ and $\kappa$ respectively. The parameter \( \Theta \) typically refers to the parameters of the GNN, while the sparsifier \( \kappa(\cdot) \), on the other hand, is crafted to take an ego graph \( \mathcal{G}^{(i)} \) and output its sparsified version \( \Tilde{\mathcal{G}^{(i)}} \), guided by a specific pruning paradigm $C$ and sparsity $s^m\%$:
\begin{equation}\label{eq:remove}
\kappa^m(\mathcal{G}^{(i)}) = \{\mathcal{V}^{(i)},\mathcal{E}^{(i)}\setminus\mathcal{E}_p^{(i)} \},\; \mathcal{E}_p^{(i)} = \operatorname{TopK}\left(-C^m(\mathcal{E}), \lceil|\mathcal{E}^{(i)}|\times s^m\%\rceil \right),
\end{equation}
where \( C^m(\cdot) \) acts as the $m$-th expert's scoring function that evaluates edge importance. We leverage long-tail gradient estimation~\citep{liu2020dynamic} to ensure the $\operatorname{TopK}(\cdot)$ operator is differentiable. Furthermore, to ensure different sparsity criteria drive the sparsifier, we implement \( C^m(\cdot) \) as follows:
\begin{equation}\label{eq:4_criteria}
\begin{aligned}
C^m(e_{ij}) = \operatorname{FFN}\left( x_i, x_j,  c(e_{ij}) \right),\; c^m(e_{ij}) \in \left\{\begin{array}{c}
\text{Degree: } \left(|\mathcal{N}(v_i) + \mathcal{N}(v_j)|\right)/2\\
\text{Jaccard Similarity: }\frac{|\mathcal{N}(v_i) \cap \mathcal{N}(v_j) |}{|\mathcal{N}(v_i) \cup \mathcal{N}(v_j)|}\\
\text{ER: }(e_i - e_j)^T\mathbf{L}^{-1}(e_i - e_j)\\
\text{Gradient Magnitude: } |\partial\mathcal{L}/\partial e_{ij}|
\end{array}\right\},
\end{aligned}
\end{equation}
where $\operatorname{FFN}(\cdot)$ is a feed-forward network, $c^m(e_{ij})$ represents the prior guidance on edge significance. By equipping different sparsifiers with various priors and sparsity levels, we can customize the most appropriate pruning strategy for each node's local scenario. In practice, we select four widely-used pruning criteria including edge degree~\citep{seo2024teddy}, Jaccard similarity~\citep{murphy1996,venu2011}, effective resistance~\citep{spielman2008graph,liu2023dspar} and gradient magnitude~\citep{wan2021edge,zhang2024graph}. Details regarding these criteria and their implementations are in \Cref{app:detail_criteria}.

\subsection{Graph Mixture on Grassmann Manifold}\label{subsec:gmgm}
After employing $k$ sparsifiers driven by different criteria and sparsity levels, we are in need of an effective mechanism to ensemble these $k$ sparse subgraphs and maximize the aggregation of their advantages. A straightforward approach is voting or averaging~\citep{sagi2018ensemble}; however, such simple merging may fail to capture the intricate relationships among multi-view graphs~\citep{kang2020multi}, potentially resulting in the loss of advantageous properties from all experts. 
Inspired by recent advances in manifold representations~\citep{dong2013clustering,bendokat2024grassmann}, we develop a subspace-based sparse graph ensembling mechanism. We first provide the definition of the Grassmann manifold~\citep{bendokat2024grassmann} as follows:

\begin{definition}[Grassmann manifold]
Grassmann manifold $Gr(n, p)$ is the space of $n$-by-$p$ matrices (e.g., $\mathbf{M}$) with
orthonormal columns, where $0\leq p \leq n$, i.e.,
\begin{equation}
\label{def:grass}
Gr(n,p) = \left\{\mathbf{M}|\mathbf{M}\in\mathbb{R}^{n\times p}, \mathbf{M}^\top\mathbf{M} = \mathbf{I}\right\}.
\end{equation}
\end{definition}
According to Grassmann manifold theory, each orthonormal matrix represents a unique subspace and thus corresponds to a distinct point on the Grassmann manifold~\citep{lin2020structure}. This applies to the eigenvector matrix of the normalized Laplacian matrix ($\mathbf{U} = \mathbf{L}[:,:p] \in \mathbb{R}^{n \times p}$), which comprises the first $p$ eigenvectors and is orthonormal~\citep{merris1995survey}, and thereby can be mapped onto the Grassmann manifold.

Consider the $k$ sparse subgraphs $\{\widetilde{\mathcal{G}}^{(i)}_{m}\}_{m=1}^{k}$, their subspace representations are $\{\mathbf{U}^{(i)}_m \in \mathbb{R}^{|\mathcal{N}(v_i)|\times p}\}_{m=1}^{k}$. We aim to identify an oracle subspace $\mathbf{U}^{(i)}$ on the Grassmann manifold, which essentially represents a graph, that serves as an informative combination of $k$ base graphs. Formally, we present the following objective function:
\begin{equation}\label{eq:grass_1}
\begin{aligned}
\min_{\mathbf{U^{(i)}}\in\mathbb{R}^{|\mathcal{N}(v_i)|\times p}} \sum_{m=1}^k \left( 
\underbrace{ \operatorname{tr}(\mathbf{U^{(i)}}^\top \mathbf{L}_m\mathbf{U}^{(i)})}_{\textrm{(1) node connectivity}} +  \underbrace{\overbrace{E^{(i)}_m}^{\text{expert score}}\cdot d^2(\mathbf{U}^{(i)},\mathbf{U}^{(i)}_m)}_{\textrm{(2) subspace distance}}\right), \operatorname{s.t. } \mathbf{U^{(i)}}^\top \mathbf{U^{(i)}} = \mathbf{I}
\end{aligned}
\end{equation}
where $\operatorname{tr}(\cdot)$ calculates the trace of matrices, $\mathbf{L}_m$ is the graph Laplacian of $\mathcal{G}^{(i)}_m$, $d^2(\mathbf{U}_1,\mathbf{U}_2)$ denotes the project distance between two subspaces~\citep{dong2013clustering}, and $E^{(i)}_m$ is the expert score for the $m$-th expert, calculated by the routing network $\Psi$, which determines which expert's subspace the combined subspace should more closely align with. In~\Cref{eq:grass_1}, the first term is designed to preserve the original node connectivity based on spectral embedding, and the second term controls that individual subspaces are
close to the final representative subspace $\mathbf{U}^{(i)}$. Using the Rayleigh-Ritz Theorem~\citep{jia2001analysis}, we provide a closed-form solution for \Cref{eq:grass_1} and obtain the graph Laplacian of the ensemble sparse graph $\widehat{\mathcal{G}^{(i)}}$ as follows:
\begin{equation}\label{eq:new_L}
\widehat{\mathbf{L}^{(i)}} = \sum_{m=1}^k\left(\mathbf{L}_m - E_m^{(i)}\cdot\mathbf{U^{(i)}}^\top \mathbf{U^{(i)}}\right).
\end{equation}
We provide detailed derivations and explanations for \Cref{eq:grass_1,eq:new_L} in \Cref{app:ensemble}. Consequently, we can reformulate the function $\text{\fontfamily{lmtt}\selectfont \textbf{ESMB}}(\cdot)$ in \Cref{eq:esmb_1} as follows:
\begin{equation}\label{eq:esmb_imple}
\text{\fontfamily{lmtt}\selectfont \textbf{ESMB}}(\{\widetilde{\mathcal{G}}^{(i)}_{m}\}_{m=1}^{k}) = \{\mathbf{D} - \widehat{\mathbf{L}^{(i)}}, \mathbf{X}^{(i)}\} = \left\{\mathbf{D} - \sum_{m=1}^k\left(\mathbf{L}_m - E_m^{(i)}\cdot\mathbf{U^{(i)}}^\top \mathbf{U^{(i)}}\right), \mathbf{X}^{(i)}\right\}.
\end{equation}
On the Grassmann manifold, the subspace ensemble effectively captures the beneficial properties of each expert's sparse graph. After obtaining the final version of each node's ego-graph, $\widehat{\mathcal{G}^{(i)}} = \{\widehat{\mathbf{A}^{(i)}},\mathbf{X}^{(i)}\}$, we conduct a post-sparsification step as the graph ensembled on the Grassmann manifold can become dense again. Specifically, we obtain the final sparsity $s^{(i)}\%$ for $v_i$ by weighting the sparsity of each expert and sparsifying $\widehat{\mathcal{G}^{(i)}}$.
\begin{equation}\label{eq:post}
\widehat{\mathcal{G}^{(i)}} \leftarrow \{\operatorname{TopK}(\widehat{\mathbf{A}^{(i)}}, |\mathcal{E}^{(i)}|\times s^{(i)}\%),\mathbf{X}^{(i)} \},\; s^{(i)}\%=\frac{1}{k}\sum_{m=1}^k s^m\%.
\end{equation}

These post-sparsified $\widehat{\mathcal{G}^{(i)}}$ are then reassembled together into $\widehat{\mathcal{G}} \leftarrow \{\widehat{\mathcal{G}^{(1)}},\widehat{\mathcal{G}^{(2)}},\cdots,\widehat{\mathcal{G}^{(|\mathcal{V}|)}}\}$. Ultimately, the sparsified graph $\widehat{\mathcal{G}}$ produced by \ourmethod can be input into any MPNN~\citep{gilmer2017neural} or graph transformer~\citep{min2022transformer} architectures for end-to-end training.

\subsection{Training and Optimization}\label{sec:optim}

\paragraph{Additional Loss Functions} Following classic MoE works~\citep{shazeer2017outrageously,wang2024graph}, we introduce an expert importance loss to prevent \ourmethod from converging to a trivial solution where only a single group of experts is consistently selected:
\begin{equation}
\operatorname{Importance}(\mathcal{V}) =\sum_{i=1}^{|\mathcal{V}|}\sum_{j=1}^{k}E^{(i)}_m,\;\;\mathcal{L}_{\text{importance}}(\mathcal{V}) = \operatorname{CV}(\operatorname{Importance}(\mathcal{V}))^2,
\end{equation}
where $\operatorname{Importance}(\mathcal{V})$ represents the sum of each node's expert scores across the node-set, $\operatorname{CV}(\cdot)$ calculates the coefficient of variation, and $\mathcal{L}_{\text{importance}}$ ensures the variation of experts. Therefore, the final loss function combines both task-specific and MoG-related losses, formulated as follows:
\begin{equation}\label{eq:final_loss}
\mathcal{L} = \mathcal{L}_{\text{task}} + \lambda\cdot\mathcal{L}_{\text{importance}},
\end{equation}
where $\lambda$ is a hand-tuned scaling factor, with its sensitivity analysis placed in \Cref{sec:sensitivity}.

\vspace{-0.5em}
\paragraph{Complexity Analysis} To better illustrate the effectiveness and clarity of \ourmethod, we provide a comprehensive {algorithmic table} in \Cref{app:algo} and detailed {complexity analysis} in \Cref{app:complexity}. To address concerns regarding the runtime efficiency of \ourmethod, we have included an empirical analysis of efficiency in \Cref{sec:exp_eff}.

\section{Experiments}
\vspace{-1em}
In this section, we conduct extensive experiments to answer the following research questions: 
(\textbf{RQ1}) Can \ourmethod effectively help GNNs combat graph sparsity?
(\textbf{RQ2})  Does \ourmethod genuinely accelerate the GNN inference?
(\textbf{RQ3}) Can \ourmethod help boost GNN performance?
(\textbf{RQ4}) How sensitive is \ourmethod to its key components and parameters?

\vspace{-0.6em}
\subsection{Experiment Setup}
\vspace{-0.6em}
\paragraph{Datasets and Backbones} We opt for four large-scale OGB benchmarks~\citep{hu2020open}, including \textsc{ogbn-arxiv}, \textsc{ogbn-proteins} and \textsc{ogbn-products} for {node classification}, and \textsc{ogbg-ppa} for {graph classification}. The dataset splits are given by \citep{hu2020open}. Additionally, we choose two superpixel datasets, MNIST and CIFAR-10~\citep{knyazev2019understanding}. We select GraphSAGE~\citep{hamilton2017inductive}, DeeperGCN~\citep{li2020deepergcn}, and PNA~\citep{corso2020principal} as the GNN backbones. More details are provided in \Cref{app:exp_details}.
\vspace{-0.7em}
\paragraph{Parameter Configurations}\label{para:param}  For {\ourmethod}, we adopt the $m=4$ sparsity criteria outlined in \Cref{sec:sparsifier}, assigning $n=3$ different sparsity levels $\{s_1, s_2, s_3\}$ to each criterion, resulting in a total of $K=m \times n=12$ experts. We select $k=2$ sparsifier experts for each node, and set the loss scaling factor $\lambda=1e-2$ across all datasets and backbones. By adjusting the sparsity combination, we can control the global sparsity of the entire graph. We present more details on parameter settings in \Cref{app:exp_param}, and a recipe for adjusting the graph sparsity in \Cref{app:adjust}.

\vspace{-0.6em}
\subsection{\ourmethod as Graph Sparsifier (RQ1 \& RQ2)}\label{sec:exp_spar}
\begin{figure}[!tpb]
\setlength{\abovecaptionskip}{6pt}
\centering
\includegraphics[width=\textwidth]{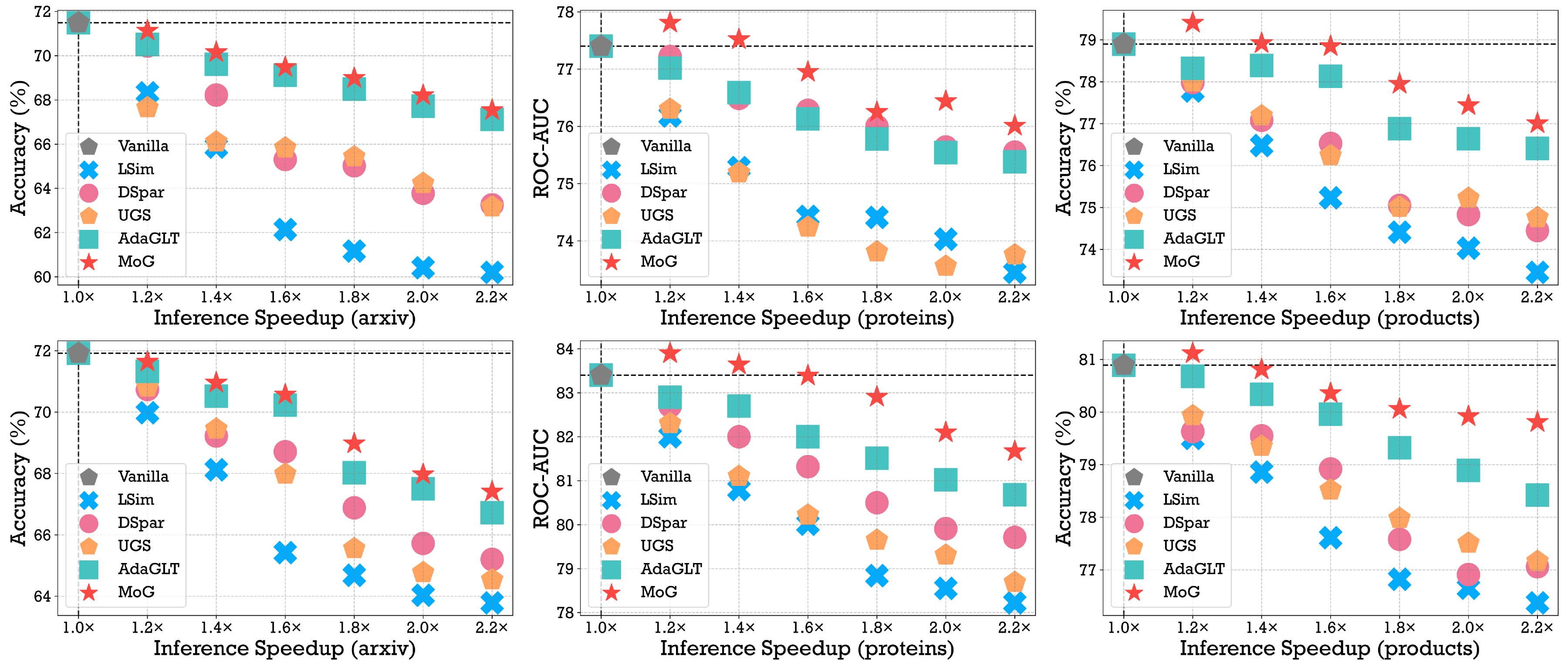}
\vspace{-1.5em}
\caption{The trade-off between inference speedup and model performance for MoG and other sparsifiers. The first and second rows represent results on GraphSAGE and DeeperGCN, respectively. The gray pentagon represents the performance of the original GNN without sparsification.} \label{fig:inference_speed}
\vspace{-1.3em}
\end{figure}

\vspace{-0.5em}
To answer \textbf{RQ1} and \textbf{RQ2}, we comprehensively compare \ourmethod with eleven widely-used topology-guided sparsifiers and five semantic-guided sparsifiers, as outlined in \Cref{tab:rq1_arxiv}, with more detailed explanations in \Cref{app:exp_sparsifier}. The quantitative results on five datasets are shown in \Cref{tab:rq1_arxiv,tab:rq1_arxiv_2,tab:rq1_products,tab:rq1_cifar,tab:rq1_proteins} and the efficiency comparison is in \Cref{fig:inference_speed}. We give the following observations (\textbf{Obs.}):

\newcommand{\blue}[1]{$_{\color{BlueGreen}\downarrow #1}$}
\newcommand{\red}[1]{$_{\color{RedOrange}\uparrow #1}$}
\begin{table*}[!tbp]
    \caption{Node classification performance comparison to state-of-the-art sparsification methods. All methods are trained using \textbf{GraphSAGE}, and the reported metrics represent the average of \textbf{five runs}. We denote methods with $\dag$ that do not have precise control over sparsity; their performance is reported around the target sparsity $\pm 2\%$. ``Sparsity \%'' refers to the ratio of removed edges as defined in \Cref{para:notation}. ``OOM'' and ``OOT'' denotes out-of-memory and out-of-time, respectively.
    }
\label{tab:rq1_arxiv}
    \centering
    \footnotesize
    \setlength{\tabcolsep}{1pt}
    \resizebox{\textwidth}{!}{
    \begin{tabular}{cc|cccc|cccc}
    \toprule
    \multirow{3}{*}{} & Dataset  & \multicolumn{4}{c|}{\textsc{ogbn-arxiv} (Accuracy$\uparrow$)} & \multicolumn{4}{c}{\textsc{ogbn-proteins} (ROC-AUC$\uparrow$)}  \\
    \midrule
    & Sparsity \% & 10  & 30 & 50 & 70 & 10  & 30 & 50 & 70 \\ \midrule
    \parbox[t]{4mm}{\multirow{9}{*}{\rotatebox[origin=c]{90}{Topology-guided}}}
    & Random
    & 70.03\blue{1.46} & 68.40\blue{3.09} & 64.32\blue{7.17} & 61.18\blue{10.3}& 76.72\blue{0.68} & 75.03\blue{2.37} & 73.58\blue{3.82} & 72.30\blue{5.10}\\
    
    & Rank Degree$^{\dag}$~\citep{voudigari2016rank}  & 68.13\blue{3.36} & 67.01\blue{4.48} & 65.58\blue{5.91} & 62.17\blue{9.32} & 77.47\red{0.07} & 76.15\blue{1.25} & 75.59\blue{1.81} & 74.23\blue{3.17}  \\
    
    & Local Degree$^{\dag}$~\citep{hamann2016structure}   & 68.94\blue{2.55} & 67.20\blue{4.29} & 65.45\blue{6.04} & 65.59\blue{5.90} & 76.20\blue{1.20} & 76.05\blue{1.35} & 76.09\blue{1.31} & 72.88\blue{4.52}\\
    
    & Forest Fire$^{\dag}$~\citep{Leskovec2006}   & 68.39\blue{3.10} & 68.10\blue{3.39} & 67.36\blue{4.13} & 65.22\blue{6.27} & 76.50\blue{0.90} & 75.37\blue{2.03} & 74.29\blue{3.11} & 72.11\blue{5.29}\\

    & G-Spar~\citep{murphy1996finley} 
   & 71.30\blue{0.19} & 69.29\blue{2.20} & 65.56\blue{5.93} & 65.49\blue{6.00} & 77.38\blue{0.02} & 77.36\blue{0.04} & 76.02\blue{1.38} & 75.89\blue{1.51}\\

    & LSim$^{\dag}$~\citep{satuluri2011local}
    & 69.22\blue{2.27} & 66.15\blue{5.34} & 61.07\blue{10.4} & 60.32\blue{11.2} & 76.83\blue{0.57} & 76.01\blue{1.39} & 74.83\blue{2.57} & 73.65\blue{3.75}\\
    
    & SCAN~\citep{xu2007scan} & 71.55\red{0.06} & 69.27\blue{2.22} & 65.14\blue{6.35} & 64.72\blue{6.77}   & 77.60\red{0.20} & 76.88\blue{0.52} & 76.19\blue{1.21} & 74.32\blue{3.08}\\
    
    & ER~\citep{spielman2008graph}& 71.63\red{0.14} & 69.48\blue{2.01} & 69.00\blue{2.49} & 67.15\blue{4.34} & \multicolumn{4}{c}{OOT} \\
    
    & DSpar~\citep{liu2023dspar}
    & 71.23\blue{0.26} & 68.50\blue{2.99} & 64.79\blue{6.70} & 63.11\blue{8.38}
    & 77.34\blue{0.06} & 77.06\blue{0.34} & 76.38\blue{1.02} & 75.49\blue{1.91}
    \\
    \midrule
    \parbox[t]{4mm}{\multirow{7}{*}{\rotatebox[origin=c]{90}{Semantic-guided}}}
   
    & UGS$^{\dag}$~\citep{chen2021unified}
    & 68.77\blue{2.72} & 66.30\blue{5.19} & 65.72\blue{5.77} & 63.10\blue{8.39} 
    & 76.80\blue{0.60} & 75.46\blue{1.94} & 73.28\blue{4.12} & 73.31\blue{4.09}\\

    & GEBT~\citep{you2022early}
       & 69.04\blue{2.45} & 65.29\blue{6.20} & 65.88\blue{5.61} & 65.62\blue{5.87} 
       & 76.30\blue{1.10} & 76.17\blue{1.23} & 74.43\blue{2.97} & 74.12\blue{3.28} \\

    & MGSpar~\citep{wan2021edge}
    & 70.22\blue{1.27} & 69.13\blue{2.36} & 68.27\blue{3.22} & 66.55\blue{4.94} 
    & \multicolumn{4}{c}{OOM} \\

    & ACE-GLT$^{\dag}$~\citep{wang2023adversarial}
    & 71.88\red{0.39} & 70.14\blue{1.35} & 68.08\blue{3.41} & 67.04\blue{4.45} 
    & 77.59\red{0.19} & 76.14\blue{1.26} & 75.43\blue{1.97} & 73.28\blue{4.12} \\
    
    & WD-GLT$^{\dag}$~\citep{hui2022rethinking}
    & 71.92\red{0.43} & 70.21\blue{1.28} & 68.30\blue{3.19} & 66.57\blue{4.92} 
    & \multicolumn{4}{c}{OOM}\\
    
    & AdaGLT~\citep{zhang2024graph}
    & 71.22\blue{0.27} & 70.18\blue{1.31} &\colorbox[HTML]{DAE8FC}{\textbf{69.13\blue{2.36}}} & 67.02\blue{4.47}  
    & 77.49\red{0.09} & 76.76\blue{1.64} & 76.00\blue{2.40} & 75.44\blue{2.96}  \\
    
    & \textbf{\ourmethod (Ours)}$^{\dag}$ & \colorbox[HTML]{DAE8FC}{\textbf{71.93\red{\textbf{0.44}}}} 
    & \colorbox[HTML]{DAE8FC}{\textbf{70.53}\blue{\textbf{0.96}}} 
    & {{69.06\blue{{2.43}}}} & \colorbox[HTML]{DAE8FC}{\textbf{67.31\blue{\textbf{4.18}}}}
    & \colorbox[HTML]{DAE8FC}{\textbf{77.78\red{\textbf{0.38}}}} 
    & \colorbox[HTML]{DAE8FC}{\textbf{77.49}\red{\textbf{0.09}}} 
    &  \colorbox[HTML]{DAE8FC}{\textbf{76.46\blue{\textbf{0.94}}}} & \colorbox[HTML]{DAE8FC}{\textbf{76.12\blue{\textbf{1.28}}}}
      \\

    \midrule
    \multicolumn{2}{c|}{Whole Dataset} & \multicolumn{4}{c|}{71.49$_{\pm0.01}$} & \multicolumn{4}{c}{77.40$_{\pm0.1}$} \\
    \bottomrule
    \end{tabular}
}
\vspace{-0.8em}
\end{table*}

\vspace{-0.8em}
\paragraph{Obs. \ding{182} \ourmethod demonstrates superior performance in both transductive and inductive settings.} As shown in \Cref{tab:rq1_arxiv,tab:rq1_graph,tab:rq1_products,tab:rq1_arxiv_2,tab:rq1_proteins}, \ourmethod outperforms other sparsifiers in both transductive and inductive settings. Specifically, for node classification tasks, \ourmethod achieves a $0.09\%$ performance improvement while sparsifying $30\%$ of the edges on \textsc{ogbn-proteins}+GraphSAGE. Even when sparsifying $50\%$ of the edges on \textsc{ogbn-proteins}+DeeperGCN, the ROC-AUC only drops by $0.81\%$. For graph classification tasks, \ourmethod can remove up to $50\%$ of the edges on \textsc{MNIST} with a $0.14\%$ performance improvement, surpassing other sparsifiers by $0.99\%\sim12.97\%$ in accuracy.
\vspace{-0.5em}
\paragraph{Obs. \ding{183} Different datasets and backbones exhibit varying sensitivities to sparsification.} As shown in \Cref{tab:rq1_arxiv,tab:rq1_arxiv_2}, despite \textsc{ogbn-proteins} being relatively insensitive to sparsification, sparsification at extremely high levels (e.g., $70\%$) causes more performance loss for GraphSAGE compared to DeeperGCN, with the former experiencing a $2.28\%$ drop and the latter only $1.07\%$, which demonstrates the varying sensitivity of different GNN backbones to sparsification. Similarly, we observe in \Cref{tab:rq1_graph} that the \textsc{MNIST} dataset shows a slight accuracy increase even with $50\%$ sparsification, whereas the \textsc{ogbg-ppa} dataset suffers a $1.86\%$ performance decline, illustrating the different sensitivities to sparsification across graph datasets.
\vspace{-0.4em}
\paragraph{Obs. \ding{184} \ourmethod can effectively accelerate GNN inference with negligible performance loss.} \Cref{fig:inference_speed} illustrates the actual acceleration effects of \ourmethod compared to other baseline sparsifiers. It is evident that \ourmethod achieves $1.6\times$ \textit{lossless acceleration} on \textsc{ogbn-proteins}+DeeperGCN and \textsc{ogbn-products}+GraphSAGE, meaning the performance is equal to or better than the vanilla backbone. Notably, on \textsc{ogbn-products}+DeeperGCN, \ourmethod achieves $3.3\times$ acceleration with less than a $1.0\%$ performance drop. Overall, \ourmethod provides significantly superior inference acceleration compared to its competitors.

\begin{table*}[!tbp]
    \caption{Graph classification performance comparison to state-of-the-art sparsification methods. The reported metrics represent the average of \textbf{five runs}. 
    }
\label{tab:rq1_graph}
     \centering
    \footnotesize
    \setlength{\tabcolsep}{1pt}
    \resizebox{\textwidth}{!}{
    \begin{tabular}{cc|cccc|cccc}
    \toprule
    \multirow{3}{*}{} & Dataset  & \multicolumn{4}{c|}{\textsc{MNIST} + PNA (Accuracy $\uparrow$)} & \multicolumn{4}{c}{\textsc{ogbn-ppa} + DeeperGCN (Accuracy $\uparrow$)}  \\
    \midrule
    & Sparsity \% & 10  & 30 & 50 & 70 & 10  & 30 & 50 & 70 \\ \midrule
    \parbox[t]{4mm}{\multirow{9}{*}{\rotatebox[origin=c]{90}{Topology-guided}}}
    & Random
    & 94.61\blue{2.74} & 87.23\blue{10.1} & 84.82\blue{12.5} & 80.07\blue{17.3}
    & 75.44\blue{1.65} & 73.81\blue{4.09} & 71.97\blue{5.12} & 69.62\blue{7.47}
    \\
    
    & Rank Degree$^{\dag}$~\citep{voudigari2016rank}  
    & 96.42\blue{0.93} & 94.23\blue{3.12} & 92.36\blue{4.99} & 89.20\blue{8.15} 
    & 75.81\blue{1.28} & 74.99\blue{2.10} & 74.12\blue{2.97} & 70.68\blue{6.41}\\
    
    & Local Degree$^{\dag}$~\citep{hamann2016structure}   
    & 95.95\blue{1.40} & 93.37\blue{3.98} & 90.11\blue{7.24} & 86.24\blue{11.1} 
    & 76.43\blue{0.66} & 75.87\blue{1.22} & 72.11\blue{4.98} & 69.93\blue{7.16}\\
    
    & Forest Fire$^{\dag}$~\citep{Leskovec2006}   
    & 96.75\blue{0.60} & 95.42\blue{1.93} & 95.03\blue{2.32} & 93.10\blue{4.25} 
    & 76.38\blue{0.71} & 75.33\blue{1.76} & 73.18\blue{3.91} & 71.49\blue{5.60}\\
    
    & G-Spar~\citep{murphy1996finley} 
    & 97.10\blue{0.25} & 96.59\blue{0.76} & 94.36\blue{2.99} & 92.48\blue{4.87} 
    & 77.68\red{0.59} & 73.90\blue{3.19} & 69.52\blue{7.57} & 68.10\blue{8.99}\\

    & LSim$^{\dag}$~\citep{satuluri2011local}
    & 95.79\blue{1.56} & 92.14\blue{5.21} & 92.29\blue{5.06} & 91.95\blue{5.40}  
    & 76.04\blue{1.05} & 74.40\blue{2.69} & 72.78\blue{4.31} & 68.21\blue{8.88}\\
    
    & SCAN~\citep{xu2007scan} 
    & 95.81\blue{1.54} & 93.48\blue{3.87} & 90.18\blue{7.17} & 86.48\blue{10.9} 
    & 75.23\blue{1.86} & 75.18\blue{1.91} & 72.48\blue{4.61} & 71.11\blue{5.98}\\
    
    & ER~\citep{spielman2008graph}
    & 94.77\blue{2.58} & 93.91\blue{3.44} & 93.45\blue{3.90} & 91.07\blue{6.28} 
    & 77.94\red{0.85} & 75.15\blue{1.94} & 73.23\blue{3.86} & 72.74\blue{4.35} \\
    
    & DSpar~\citep{liu2023dspar}
    & 94.97\blue{2.38} & 93.80\blue{3.55} & 92.23\blue{5.12} & 90.48\blue{6.87} 
    & 76.33\blue{0.76} & 73.37\blue{3.72} & 72.98\blue{4.11} & 70.77\blue{6.32}\\
    \midrule
    
    \parbox[t]{4mm}{\multirow{2}{*}{\rotatebox[origin=c]{90}{Semantic}}}
    & ICPG~\citep{sui2021inductive}
    & 97.69\red{0.34} & 97.39\red{0.04} & 96.80\blue{0.55} & 93.77\blue{3.58} 
    & 77.36\red{0.27} & 75.24\blue{1.85} & 73.18\blue{3.91} & 71.09\blue{6.00}\\

    & AdaGLT~\citep{zhang2024graph}
    & 97.31\blue{0.04} & 96.58\blue{0.77} & 94.14\blue{3.21} & 92.08\blue{5.27}  
    & 76.22\blue{0.87} & 73.54\blue{3.55} & 70.10\blue{6.99} & 69.28\blue{7.81}  \\
    
    & \textbf{\ourmethod (Ours)}$^{\dag}$ & \colorbox[HTML]{DAE8FC}{\textbf{97.80\red{\textbf{0.45}}}} 
    & \colorbox[HTML]{DAE8FC}{\textbf{97.74}\red{\textbf{0.39}}} 
    & \colorbox[HTML]{DAE8FC}{\textbf{97.79}\red{\textbf{0.44}}}  & \colorbox[HTML]{DAE8FC}{\textbf{95.30\blue{\textbf{2.05}}}}
    & \colorbox[HTML]{DAE8FC}{\textbf{78.43\red{\textbf{1.34}}}} 
    & \colorbox[HTML]{DAE8FC}{\textbf{77.90}\red{\textbf{0.81}}} 
    & \colorbox[HTML]{DAE8FC}{\textbf{75.23}\blue{\textbf{1.86}}}  & \colorbox[HTML]{DAE8FC}{\textbf{73.09\blue{\textbf{4.00}}}}
      \\

    \midrule
    \multicolumn{2}{c|}{Whole Dataset} & \multicolumn{4}{c|}{97.35$_{\pm0.07}$} & \multicolumn{4}{c}{77.09$_{\pm0.04}$} \\
    \bottomrule
    \end{tabular}
}
\vspace{-0.8em}
\end{table*}

\vspace{-1em}
\subsection{\ourmethod as Performance Booster (RQ3)}\label{sec:exp_boost}
\vspace{-0.7em}

In the context of \textbf{RQ3}, \ourmethod is developed to augment GNN performance by selectively removing a limited amount of noisy and detrimental edges, while simultaneously preventing excessive sparsification that could degrade GNN performance. Consequently, we uniformly set the sparsity combination to $\{90\%, 85\%, 80\%\}$. We combine \ourmethod with state-of-the-art GNNs on both node-level and graph-level tasks. The former include RevGNN~\citep{li2021training} and GAT+BoT~\citep{wang2021bag}, which rank fourth and seventh, respectively, on the \textsc{ogbn-proteins} benchmark, and the latter include PAS~\citep{wei2021pooling} and DeeperGCN~\citep{li2020deepergcn}, ranking fourth and sixth on the \textsc{ogbn-ppa} benchmark. We observe from \Cref{tab:rq3}:
\vspace{-0.8em}
\paragraph{Obs. \ding{185} \ourmethod can assist the ``top-student'' backbones to learn better.} Despite RevGNN and PAS being high-ranking backbones for \textsc{ogbn-proteins} and \textsc{ogbg-ppa}, \ourmethod still achieves non-marginal performance improvements through moderate graph sparsification: $1.02\%\uparrow$ on RevGNN+\textsc{ogbn-proteins} and $1.74\%\uparrow$ on DeeperGCN+\textsc{ogbg-ppa}. This demonstrates that \ourmethod can effectively serve as a plugin to boost GNN performance by setting a relatively low sparsification rate.


\begin{table}[!tpb]
\caption{Node classification results on \textsc{ogbn-proteins} with RevGNN and GAT+BoT and graph classification results on \textsc{ogbg-ppa} with PAS and DeeperGCN.  Mean and standard deviation values from \textbf{five} random runs are presented.}\label{tab:rq3}
\vspace{-0.5em}
\begin{center}
\begin{tabular}{l|cc|cc}
\toprule
&\multicolumn{2}{c|}{\textsc{ogbn-proteins} (ROC-AUC$\uparrow$)} & \multicolumn{2}{c}{\textsc{ogbg-ppa} (Accuracy$\uparrow$)}\\
\midrule
{Model} & \makecell{RevGNN} & \makecell{GAT+BoT}  & \makecell{PAS} & \makecell{DeeperGCN}  \\
\midrule
w/o \ourmethod      & 88.14$\pm$ 0.24 & 88.09	$\pm$ 0.16 & 78.28 $\pm$ 0.24 & 77.09	$\pm$ 0.04
   \\
\midrule
\multirow{2}{*}{w/ \ourmethod} & \textbf{89.04} $\pm$ 0.72 &  \textbf{88.72} $\pm$ 0.50 & \textbf{78.66} $\pm$ 0.47 & \textbf{78.43} $\pm$ 0.19\\
& (Sparsity: 9.2\%)  & (Sparsity: 12.7\%) & (Sparsity: 6.6\%)  & (Sparsity: 10.8\%) \\
\bottomrule
\end{tabular}
\end{center}
\vspace{-1em}
\end{table}
\vspace{-0.6em}

\subsection{Sensitivity Analysis (RQ4)}\label{sec:sensitivity}
\vspace{-0.5em}
To answer \textbf{RQ4}, we perform a sensitivity analysis on the two most important parameters in \ourmethod: the number of selected experts \( k \) and the expert importance loss coefficient \(\lambda\). We compared the performance of \ourmethod when choosing different numbers of experts per node, as outlined in \Cref{fig:abla_k}. The effect of different scaling factors \(\lambda\) on \textsc{ogbn-proteins}+DeeperGCN is shown in \Cref{tab:abla_lambda}. Based on the results of the above sensitivity analysis, we observe that:

\vspace{-0.5em}
\paragraph{Obs. \ding{186} Sparse expert selection helps customized sparsification.}
It can be observed FROM \Cref{fig:abla_k} that the optimal \( k \) varies with the level of graph sparsity. At lower sparsity (10\%), \( k=1 \) yields relatively good performance. However, as sparsity increases to 50\%, model performance peaks at \( k=4 \), suggesting that in high sparsity environments, more expert opinions contribute to better sparsification. Notably, when \( k \) increases to 6, \ourmethod's performance declines, indicating that a more selective approach in sparse expert selection aids in better model generalization. For a balanced consideration of performance and computational efficiency, we set \( k=2 \) in all experiments. We further provide sensitivity analysis results of paramter $k$ on more datasets, as shown in \Cref{app:sen_proteins_rev}.
\vspace{-0.9em}
\paragraph{Obs. \ding{187} Sparsifier load balancing is essential.} We conduct a sensitivity analysis of the expert importance loss coefficient \(\lambda\). A larger \(\lambda\) indicates greater variation in the selected experts. As shown in \Cref{tab:abla_lambda}, \(\lambda=0\) consistently resulted in the lowest performance, as failing to explicitly enforce variation among experts leads to the model converging to a trivial solution with the same set of experts~\citep{wang2024graph,shazeer2017outrageously}. Conversely, \(\lambda=1e-1\) performed slightly better than \(\lambda=1e-2\) at higher sparsity levels, supporting the findings in Obs. 5 that higher sparsity requires more diverse sparsifier experts.
\vspace{0.3em}

\begin{minipage}[!t]{\linewidth}
    \begin{minipage}[!t]{0.58\linewidth}
    \centering
    \includegraphics[width=.95\linewidth]{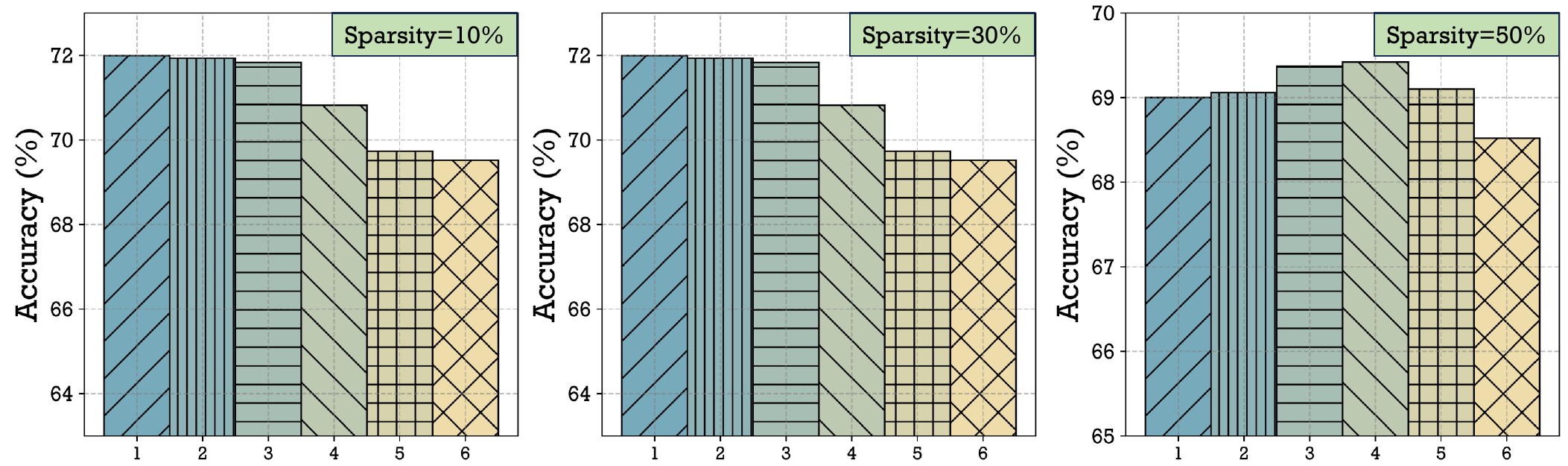}
 \vspace{-0.5em} \makeatletter\def\@captype{figure}\makeatother\caption{Sensitivity study on parameter $k$, \textit{i.e.}, how many experts are chosen per node. The results are reported based on \textsc{ogbn-arxiv}+GraphSAGE. }\label{fig:abla_k}
    \end{minipage}
    \hspace{0.1em}
    \begin{minipage}[!t]{0.40\linewidth}
    \centering
		\renewcommand{\arraystretch}{1.1}
    \scriptsize
    \tabcolsep=0.85mm
    \begin{tabular}{c|cccc}
    \toprule
    $\lambda$ & 0 & 1e-2 & 1e-1\\
    \midrule
    $10\%$ & $81.19_{\pm 0.08}$  & $\mathbf{83.32}_{\pm 0.19}$ & $83.04_{\pm 0.23}$  \\
    $30\%$  & $79.77_{\pm 0.08}$  & $\mathbf{82.14}_{\pm 0.23}$ & $82.08_{\pm 0.25}$ \\
    $50\%$ & $79.40_{\pm 0.06}$  & $81.79_{\pm 0.21}$ & $\mathbf{82.04}_{\pm 0.20}$ \\
    $70\%$ & $78.22_{\pm 0.13}$  & ${80.90}_{\pm 0.24}$ & $\mathbf{80.97}_{\pm 0.28}$\\
    \bottomrule
    \end{tabular}%
    \makeatletter\def\@captype{table}\makeatother\caption{Sensitivity study on scaling factor $\lambda$. The results are reported on \textsc{ogbn-proteins}+DeeperGCN.}\label{tab:abla_lambda}
    \end{minipage}
\end{minipage}

\begin{table}[!htpb]
\vspace{-1em}
\caption{Running time efficiency comparison on \textsc{ogbn-products}+GraphSAGE. We consistently set $n=4, k=2$, corresponding to utilizing 4 pruning criteria and selecting 2 experts for each node, and vary $m \in \{1,2,3\}$ to check how the training cost grows with $m$ increasing. }\label{tab:running_efficiency}
\vspace{0.1em}
\centering
{\small
\begin{tabular}{c|cc|cc}
\toprule
\makecell{Sparsity} &\multicolumn{2}{c|}{30\%} & \multicolumn{2}{c}{50\%}\\
\midrule
\makecell{Metric}  &  Per-epoch Time (s) & Accuracy (\%)  &  Per-epoch Time (s)  & Accuracy (\%) \\
\midrule
\makecell{Random}    & 18.71$\pm$ 0.14 & 74.21$\pm$ 0.28 & 15.42 $\pm$ 0.24 & 71.08$\pm$ 0.34
   \\
\makecell{AdaGLT} & 23.55$\pm$ 0.20 & 77.30 $\pm$ 0.54 & 21.68 $\pm$ 0.26 & 74.38 $\pm$ 0.79\\
\midrule

\makecell{\ourmethod {\scriptsize($m=1,K=3$)}} & 20.18$\pm$ 0.14 &  {77.75} $\pm$ 0.22 & 18.19 $\pm$ 0.30 & 76.10 $\pm$ 0.49\\

\makecell{\ourmethod {\scriptsize($m=2,K=6$)}} & 21.25$\pm$ 0.22 &  {78.23} $\pm$ 0.29 & 19.70 $\pm$ 0.30 & {76.43} $\pm$ 0.49\\

\makecell{\ourmethod {\scriptsize($m=3,K=12$)}} & 23.19$\pm$ 0.18 &  {78.15} $\pm$ 0.32 & 20.83 $\pm$ 0.29 & {76.98} $\pm$ 0.49\\
\bottomrule
\end{tabular}}
\vspace{-0.2em}
\end{table}

\vspace{-0.3em}
\subsection{Efficiency Analysis and Ablation Study (RQ4)}\label{sec:exp_eff}
\vspace{-0.5em}
\paragraph{Efficiency Analysis} To verify that \ourmethod can achieve better results with less additional training cost than the previous SOTA methods, we compare the accuracy and the time efficiency of \ourmethod with AdaGLT on \textsc{ogbn-product}+GraphSAGE, as outlined in \Cref{tab:running_efficiency}. We have:
\vspace{-0.8em}
\paragraph{Obs. \ding{188} \ourmethod can achieve better accuracy with less additional training cost.} It is evident in \Cref{tab:running_efficiency} that \ourmethod incurs less additional training cost compared to AdaGLT while achieving significant improvements in sparsification performance. More importantly, we demonstrate that with $k=2$, \ourmethod does not incur significantly heavier training burdens as the number of sparsifiers increases. Specifically, at $s\%=50\%$, the difference in per epoch time between \ourmethod ($K=3$) and \ourmethod ($K=12$) is only $2.63$ seconds, consistent with the findings of mainstream sparse MoE approaches~\citep{wang2024graph}.
\vspace{-0.4em}
\paragraph{Ablation Study} We test three different settings of $\epsilon$ (in \Cref{eq:routing}) on \textsc{Ogbn-Arxiv}+GraphSAGE: (1) $\epsilon\sim\mathcal{N}(0,\mathbf{I})$, (2) $\epsilon=0$, and (3) $\epsilon=0.2$, presented in \Cref{tab:ablation}. Our key finding is that randomness in gating networks consistently benefits our model. More results and detailed analysis can be found in \Cref{app:ablation}.

\vspace{-0.5em}
\section{Conclusion \& Limitation}\label{sec:conclusion}
\vspace{-0.5em}
In this paper, we introduce a new graph sparsification paradigm termed \ourmethod, which leverages multiple graph sparsifiers, each equipped with distinct sparsity levels and pruning criteria. \ourmethod selects the most suitable sparsifier expert based on each node's local context, providing a customized graph sparsification solution, followed by an effective mixture mechanism on the Grassmann manifold to ensemble the sparse graphs produced by various experts.  Extensive experiments on four large-scale OGB datasets and two superpixel datasets have rigorously demonstrated the effectiveness of \ourmethod. A potential limitation of \ourmethod is its current reliance on 1-hop decomposition to represent each node's local context. The performance of extending this approach to $k$-hop contexts remains unexplored, suggesting a possible direction for future research.

\bibliography{iclr2025_conference}
\bibliographystyle{iclr2025_conference}

\appendix

\section{Notations}
We conclude the commonly used notations throughout the manuscript in \Cref{tab:notation}.
\begin{table}[!htbp]\footnotesize
  \centering
  \caption{The notations that are commonly used in the manuscript.}
   \setlength{\tabcolsep}{16pt} 
   \renewcommand\arraystretch{1.3} %
  \vspace{1em}
    \begin{tabular}{cc}
    \toprule
    Notation & Definition \\
    \midrule
    $\mathcal{G} = \left\{\mathcal{V}, \mathcal{E} \right\} = \left\{  \mathbf{A},\mathbf{X}\right\}$          & Input graph \\
    $\mathbf{A}$  & Input adjacency matrix \\
    $\mathbf{X}$  & Node feature matrix \\
    $\mathbf{L}$ & Graph Laplacian matrix\\
    $\texttt{COMB}(\cdot)$ & GNN ego-node transformation function\\
    $\texttt{AGGR}(\cdot)$ & GNN message aggregation function\\
    $\texttt{ESMB}(\cdot)$ & Sparse graph combination function\\
    $s\%$ & Sparsity ratio (the ratio of removed edges)\\
    $v_i$ & The $i$-th node in $\mathcal{G}$\\
    $x_i$ & Node feature vector for $v_i$\\
    $\mathcal{G}^{(i)}$ & The 1-hop ego-graph for $v_i$\\
    $\phi(\mathcal{G}^{(i)})$ & Routing network\\
    $K$ & The number of total sparsifier experts\\
    $k$ & The number of selected sparsifier experts per node\\
    $W_g, W_n$ & Trainable parameters in the routing network\\
    $\kappa(\mathcal{G})$ & A graph sparsifier\\
    $\mathcal{G}^{(i)}_m$ &  The sparse ego-graph of $v_i$ produced by the $m$-th graph sparsifier\\
    $\widehat{\mathcal{G}^{(i)}} = \{ \widehat{\mathcal{V}^{(i)}}, \widehat{\mathcal{E}^{(i)}}\}$ & The ensemble sparse graph produced by \ourmethod for $v_i$\\
    $\mathcal{E}_p^{(i)}$ & Edges removed surrounding $v_i$\\
    $c^m(e_{ij})$ & Prior guidance on edge importance $e_{ij}$\\

    \bottomrule
    \end{tabular}%
  \label{tab:notation}%
\end{table}%

\section{Deatils on Pruning Criteria}\label{app:detail_criteria}

In this section, we will thoroughly explain the four pruning criteria we selected and the rationale behind these choices.
\begin{itemize}[leftmargin=*]
    \item \textbf{Edge degree} of $e_{ij}$ is defined as follows:

    \begin{equation}
        \operatorname{Degree}\;(e_{ij}) = \frac{1}{2}\left( | \mathcal{N}(v_i) + \mathcal{N}(v_j) | \right).
    \end{equation}
Previous methods~\citep{wang2022pruning,seo2024teddy} have explicitly or implicitly used edge degree for graph sparsification. Intuitively, edges with higher degrees are more replaceable. \citep{wang2022pruning} further formalizes this intuition from the perspective of bridge edges.

    \item \textbf{Jaccard Similarity}~\citep{murphy1996finley} measures the similarity between two sets by computing the portion of shared neighbors between two nodes ($v_i$and $v_j$), as defined below:

    \begin{equation}
        \operatorname{JaccradSimilarity}\;v_i,v_j) = \frac{|\mathcal{N}(v_i) \cap \mathcal{N}(v_j) |}{|\mathcal{N}(v_i) \cup \mathcal{N}(v_j)|}.
    \end{equation}

    Jaccard similarity is widely used for its capacity for detecting clusters, hubs, and outliers on social networks~\citep{murphy1996finley,xu2007scan,satuluri2011local}.

    \item \textbf{Effective Resistance}, derived from the analogy to electrical circuits, is applied to graphs where edges represent resistors. The effective resistance of an edge is defined as the potential difference generated when a unit current is introduced at one vertex and withdrawn from the other. Once the effective resistance is calculated, a sparsified subgraph can be constructed by selecting edges with probabilities proportional to their effective resistances. Notably, \citep{spielman2008graph} proved that the quadratic form of the Laplacian for such sparsified graphs closely approximates that of the original graph. Consequently, the following inequality holds for the sparsified subgraph with high probability:
    \begin{equation}
        \forall \boldsymbol{x}\in \mathbb{R}^{|\mathcal{V}|} \ \ \ (1-\epsilon)\boldsymbol{x}^T\mathbf{L} \boldsymbol{x}\le \boldsymbol{x}^T\Tilde{\mathbf{L}}\boldsymbol{x} \le(1+\epsilon)\boldsymbol{x}^T\mathbf{L} \boldsymbol{x},
    \end{equation}
    where $\Tilde{\mathbf{L}}$ is the Laplacian of the sparsified graph, and $\epsilon > 0$ is a small number. The insight is that effective resistance reflects the significance of an edge. Effective resistance aims to preserve the quadratic form of the graph Laplacian. This property makes it suitable for applications relying on the quadratic form of the graph Laplacian, such as min-cut/max-flow problems. For computation simplicity, we do not directly utilize the definition of effective resistance, and use its approximation version~\citep{liu2023dspar}.

    \item \textbf{Gradient Magnitude}, a widely used pruning criterion, is prevalent not only in the field of graph sparsification but also in classical neural network pruning. Numerous studies~\citep{lee2018snip,tessera2021keep,dettmers2019sparse} leverage gradient magnitude to estimate parameter importance. Specifically for graph sparsification, MGSpar~\citep{wan2021edge} was the first to propose using meta-gradient to estimate edge importance. We consider gradient magnitude a crucial indicator of the graph's topological structure during training. Therefore, we explicitly design some sparsifier experts to focus on this information.
    
\end{itemize}

\section{Graph Mixture on Grassmann Manifold}\label{app:ensemble}

In this section, we detail how we leverage the concept of Grassmann Manifold to effectively combine different sparse (ego-)graphs output by various sparsifiers. 

According to \Cref{def:grass}, each orthonormal matrix represents a unique subspace and thus corresponds to a distinct point on the Grassmann manifold~\citep{lin2020structure}. This applies to the eigenvector matrix of the normalized Laplacian matrix ($\mathbf{U} = \mathbf{L}[:,:p] \in \mathbb{R}^{n \times p}$), which comprises the first $p$ eigenvectors and is orthonormal~\citep{merris1995survey}, and thereby can be mapped onto the Grassmann manifold. Additionally, each row of the eigenvector matrix encapsulates the spectral embedding of each node in a p-dimensional space, where adjacent nodes have similar embedding vectors. This subspace representation, summarizing graph information, is applicable to various tasks such as clustering, classification, and graph merging~\citep{dong2013clustering}.

In the context of \ourmethod, we aim to efficiently find the final version that aggregates all the excellent properties of each point's $k$ versions of sparse ego-graph $\{\widetilde{\mathcal{G}}^{(i)}_{m}\}_{m=1}^{k}$ on the Grassmann Manifold. Moreover, this should guided by the expert scores computed by the routing network in \Cref{sec:routing}. Let $\mathbf{D}_{m}$ and $\mathbf{A}_m$ denote the degree matrix and the adjacency matrix for $\widetilde{\mathcal{G}}^{(i)}_m$ (we omit the superscript $(\cdot)^{(i)}$ denoting $v_i$ for simplicity in the subsequent expressions), then the normalized graph Laplacian is defined as:
\begin{equation}
\mathbf{L}_m = \mathbf{D}_m^{-\frac{1}{2}}(\mathbf{D}_m - \mathbf{A}_m)\mathbf{D}_m^{\frac{1}{2}}.
\end{equation}
Given the graph Laplacian $\mathbf{L}_m$ for each sparse graph, we calculate the spectral embedding matrix $\mathbf{U}_m$ through trace minimization:
\begin{equation}\label{eq:obj_1}
\min_{\mathbf{U}_m\in\mathbb{R}^{|\mathcal{N}(v_m)|\times p}} \operatorname{tr}\;(\mathbf{U}_m^\top\mathbf{L}_m\mathbf{U}_m),\;\;\operatorname{s.t.}\;\mathbf{U}_m^\top\mathbf{U}_m=\mathbf{I},
\end{equation}
which can be solved by the Rayleigh-Ritz theorem. As mentioned above, each point on the Grassmann manifold can be represented by an orthonormal matrix $\mathbf{Y}\in\mathbb{R}^{|\mathcal{N}(v_i)|\times p}$ whose columns span the corresponding $p$-dimensional subspace in $\mathbb{R}^{|\mathcal{N}(v_i)|\times p}$. The distance between such subspaces can be computed as a set of principal angles $\{\theta_i\}_{i=1}^k$ between these subspaces. \citep{dong2013clustering} showed that the projection distance between two subspaces $\mathbf{Y}_1$ and $\mathbf{Y}_2$ can be represented as a separate trace minimization problem:
\begin{equation}
d^2_{\text{proj}}(\mathbf{Y}_1, \mathbf{Y}_2) = \sum_{i=1}^{p}\sin^2{\theta_i} = k - \operatorname{tr}\;(\mathbf{Y}_1\mathbf{Y}_1^\top\mathbf{Y}_2\mathbf{Y}_2^\top).
\end{equation}
Based on this, we further define the projection of the final representative subspace $\mathbf{U}$ and the $k$ sparse candidate subspace $\{\mathbf{U}_m\}_{m=1}^k$:
\begin{equation}\label{eq:23}
    d^2_{\text{proj}}(\mathbf{U}, \{\mathbf{U}_m\}_{m=1}^k) = \sum_{m=1}^{k}d^2_{\text{proj}}(\mathbf{U},\mathbf{U}_m) = p\times k - \sum_{m=1}^k\operatorname{tr}\;(\mathbf{U}\mathbf{U}^\top\mathbf{U}_m\mathbf{U}_m^\top),
\end{equation}
which ensures that individual subspaces are close to the final representative subspace $\mathbf{U}$. 

Finally, to maintain the original vertex connectivity from all $k$ sparse ego-graphs and emphasize the connectivity relationship from more reliable sparsifiers (with higher expert scores), we propose the following objective function:
\begin{equation}\label{eq:obj_2}
\min_{\mathbf{U}_m\in\mathbb{R}^{|\mathcal{N}(v_m)|\times p}} \sum_{m=1}^{k}E^{(i)}_m\left(p\times k - \sum_{m=1}^k\operatorname{tr}\;(\mathbf{U}\mathbf{U}^\top\mathbf{U}_m\mathbf{U}_m^\top)\right),
\end{equation}
where $E^{(i)}_m$ represents the expert score of the node $v_i$'s $m$-th sparsifier expert. Based on \Cref{eq:obj_1,eq:obj_2}, we present the overall objective:
\begin{equation}\label{eq:obj_final_2}
\min_{\mathbf{U^{(i)}}\in\mathbb{R}^{|\mathcal{N}(v_i)|\times p}} \sum_{m=1}^k \left( 
\underbrace{ \operatorname{tr}(\mathbf{U^{(i)}}^\top \mathbf{L}_m\mathbf{U}^{(i)})}_{\textrm{(1) node connectivity}} +  \underbrace{E^{(i)}_m\cdot d^2(\mathbf{U}^{(i)},\mathbf{U}^{(i)}_m)}_{\textrm{(2) subspace distance}}\right), \operatorname{s.t. } \mathbf{U^{(i)}}^\top \mathbf{U^{(i)}} = \mathbf{I}.
\end{equation}
For simplicity, we omit the superscript $^{(i)}$ in the following content. Substituting \Cref{eq:23} into \Cref{eq:obj_final_2}, we obtain:
\begin{equation}
\min_{\mathbf{U}}\sum_{m=1}^{k} \text{tr}(\mathbf{U}^T\mathbf{L}_m\mathbf{U}) + E_m\cdot \left(p\times k -\sum_{m=1}^k\text{tr}(\mathbf{U}\mathbf{U}^T\mathbf{U}_m\mathbf{U}_m^T)\right), \text{s.t.} \mathbf{U}^T\mathbf{U}=\mathbf{I}. 
\end{equation}
Further simplification by neglecting constant terms like $E_m\times p \times k$ yields:
\begin{equation}
\min_{\mathbf{U}}\sum_{m=1}^{k} \text{tr}(\mathbf{U}^T\mathbf{L}_m\mathbf{U}) - E_m \cdot \sum_{m=1}^k\text{tr}(\mathbf{U}\mathbf{U}^T\mathbf{U}_m\mathbf{U}_m^T), \text{s.t.} \mathbf{U}^T\mathbf{U}=\mathbf{I}.
\end{equation}
Reorganizing the trace form of the second term, we obtain:
\begin{equation}
\min_{\mathbf{U}} \text{tr} \left[ \mathbf{U}^T (\sum_{k=1}^M \mathbf{L}_m - E_m\sum_{k=1}^M\mathbf{U}_m\mathbf{U}_m^T)\mathbf{U} \right], \text{s.t.} \mathbf{U}^T\mathbf{U}=\mathbf{I}. 
\end{equation}
At this point, the optimization problem essentially becomes a trace minimization problem, and thus the solution to this minimization problem is essentially the term between $\mathbf{U}^T$ and $\mathbf{U}$, which is:
\begin{equation}
    \widehat{\mathbf{L}} = (\sum_{k=1}^M \mathbf{L}_m - E_m\sum_{k=1}^M\mathbf{U}_m\mathbf{U}_m^T) = \sum_{k=1}^M(\mathbf{L}_m - E_m\cdot \mathbf{U}_m\mathbf{U}_m^T).
\end{equation}
Since computations involving the Grassmann manifold unavoidably entail eigenvalue decomposition, concerns about computational complexity may arise. However, given that \ourmethod only operates mixtures on the ego-graph of each node, such computational burden is entirely acceptable. Specific complexity analyses are presented in \Cref{app:complexity}.

\section{Algorithm Workflow}\label{app:algo}

The algorithm framework is presented in Algo.~\ref{alg:algo}.
\begin{algorithm}[!t]
\caption{Algorithm workflow of \ourmethod}\label{alg:algo}
\Input{$\mathcal{G}=(\mathbf{A},\mathbf{X})$, GNN model $f(\mathcal{G}, \mathbf{\Theta})$, , epoch number $Q$.}
                
\Output{Sparse graph $\mathcal{G}^{\text{sub}} = \{ \mathcal{V},\mathcal{E}' \} $}

\For{\rm{iteration} $q \leftarrow 1$ \KwTo $Q$}{

\tcc{\textcolor{blue}{Ego-graph decomposition}}
Decompose $\mathcal{G}$ into ego-graph representations $\{\mathcal{G}^{(1)}, \mathcal{G}^{(2)}, \cdots, \mathcal{G}^{(N)}\}$.

\tcc{\textcolor{blue}{Sparsifier expert allocation}}
\For{\rm{node} $i \leftarrow 1$ \KwTo $|\mathcal{V}|$}{

Calculate the total $K$ expert score of $v_i$ by routing network $\psi(x_i)$\Comment*[r]{\textcolor{blue}{Eq.~\ref{eq:routing}}}

Select $k$ sparsifier expert for node $v_i$ by $\operatorname{Softmax}(\operatorname{TopK}(\psi(x_i),k))$\Comment*[r]{\textcolor{blue}{Eq.~\ref{eq:routing}}}

}

\tcc{\textcolor{blue}{Produce sparse graph condidates}}
\For{\rm{iteration} $i \leftarrow 1$ \KwTo $|\mathcal{V}|$}{
\For{\rm{sparsifier index} $m \leftarrow 1$ \KwTo $m$}{

Sparisifier $\kappa^m$ determines which edges to remove by $\mathcal{E}_p^{(i)} = \operatorname{TopK}\left(-C^m(\mathcal{E}), \lceil|\mathcal{E}^{(i)}|\times s\%\rceil \right)$\Comment*[r]{\textcolor{blue}{Eq.~\ref{eq:remove}}}

Produce sparse graph candidate $\widetilde{\mathcal{G}}^{(i)}=\kappa^m(\mathcal{G}^{(i)}) = \{\mathcal{V}^{(i)},\mathcal{E}^{(i)}\setminus\mathcal{E}_p^{(i)} \}$.

}

\tcc{\textcolor{blue}{Ensenmble sparse graphs on Grassmann manifold}}

Calculate the ensemble graph's graph Laplacian by $\widehat{\mathbf{L}^{(i)}} = \sum_{m=1}^k\left(\mathbf{L}_m - E_m^{(i)}\cdot\mathbf{U^{(i)}}^\top \mathbf{U^{(i)}}\right)$\Comment*[r]{\textcolor{blue}{Eq.~\ref{eq:new_L}}}

Obtain $v_i$'s final sparse graph by $\texttt{ESMB}(\{\widehat{\mathcal{G}^{(i)}}= \{\mathbf{D} - \widehat{\mathbf{L}^{(i)}}, \mathbf{X}^{(i)}\})$\Comment*[r]{\textcolor{blue}{Eq.~\ref{eq:esmb_imple}}}

Compute $v_i$'s weighted sparsity by $s^{(i)}\%=\frac{1}{k}\sum_{m=1}^k s^m\%$\Comment*[r]{\textcolor{blue}{Eq.~\ref{eq:post}}}

Post-sparsify $\widehat{\mathcal{G}^{(i)}}$: $\widehat{\mathcal{G}^{(i)}} \leftarrow \{\operatorname{TopK}(\widehat{\mathbf{A}^{(i)}}, |\mathcal{E}^{(i)}|\times s^{(i)}\%),\mathbf{X}^{(i)} \}$\Comment*[r]{\textcolor{blue}{Eq.~\ref{eq:post}}}
}

\tcc{\textcolor{blue}{Combine ego-graphs}}
$\widehat{\mathcal{G}} \leftarrow \{\widehat{\mathcal{G}^{(1)}},\widehat{\mathcal{G}^{(2)}},\cdots,\widehat{\mathcal{G}^{(|\mathcal{V}|)}}\}$

\tcc{\textcolor{blue}{Standard GNN training}}
Feed the sparse graph $\widehat{\mathcal{G}}$ into GNN model for any kinds of downstream training \Comment*[r]{\textcolor{blue}{Eq.~\ref{eq:gnn2}}}

Compute loss $\mathcal{L}_{\text{task}} + \lambda\cdot\mathcal{L}_{\text{importance}}$\Comment*[r]{\textcolor{blue}{Eq.~\ref{eq:final_loss}}}

Backpropagate to update GNN $f(\mathcal{G},\mathbf{\Theta})$, routing network $\psi$ and sparsifiers $\{\kappa^m\}_{m=1}^K$.

}

\end{algorithm}

\section{Complexity Analysis}\label{app:complexity}

In this section, we delve into a comprehensive analysis of the time and space complexity of \ourmethod. Without loss of generality, we consider the scenario where \ourmethod is applied to vanilla GCN. It is worth recalling that the forward time complexity of vanilla GCN is given by:
\begin{equation}
\mathcal{O}(L\times |\mathcal{E}|\times D + L \times  |\mathcal{V}| \times D^2),
\end{equation}
where $L$ is the number of GNN layers, $|\mathcal{E}|$ and $|\mathcal{V}|$ denotes the number of edges and nodes, respectively, and $D$ is the hidden dimension. Similarly, the forward space complexity of GCN is:
\begin{equation}
\mathcal{O}(L\times |\mathcal{E}| + L \times D^2 + L \times  |\mathcal{V}| \times D)
\end{equation}
When \ourmethod is applied to GCN, each sparsifier expert $\kappa(\cdot)$ essentially introduces additional complexity equivalent to that of an $\operatorname{FFN}(\cdot)$, as depicted in \Cref{eq:4_criteria}.  Incorporating the Sparse MoE-style structure, the forward time complexity of GCN+\ourmethod becomes:
\begin{equation}
\mathcal{O}(L\times |\mathcal{E}|\times D + L \times  |\mathcal{V}| \times D^2 + k\times|\mathcal{E}| \times D \times D^s),
\end{equation}
where $D^s$ represents the hidden dimension of the feed-forward network in \Cref{eq:4_criteria} and $k$ denotes the number of selected experts. Similarly, the forward space complexity is increased to:
\begin{equation}
\mathcal{O}(L\times |\mathcal{E}| + L \times D^2 + L \times  |\mathcal{V}| \times D +  k\times|\mathcal{E}| \times D \times D^s).
\end{equation}
It is noteworthy that we omit the analysis for the routing network, as its computational cost is meanwhile negligible compared to the cost of selected experts, since both $W_g\in\mathbb{R}^{K\times F}$ and $W_n\in\mathbb{R}^{K\times F}$ is in a much smaller dimension that the weight matrix $W\in\mathbb{R}^{F\times F}$ in GCN.

Furthermore, we present the additional complexity introduced by the step of graph mixture on the Grassmann manifold. For each center node's $k$ sparse ego-graphs, we need to compute the graph Laplacian and the eigenvector matrix, which incurs an extra time complexity of $\mathcal{O}(k\times (\frac{|\mathcal{E}|}{|\mathcal{V}|})^3 )$; to compute the Laplacian $\widehat{\mathbf{L}^{(i)}}$ of the final ensemble sparse graph, an additional complexity of $\mathcal{O}(k\times (\frac{|\mathcal{E}|}{|\mathcal{V}|})^2 \times p)$ is required. In the end, the complete time complexity of \ourmethod is expressed as:
\begin{equation}
\mathcal{O}\left(\underbrace{L\times |\mathcal{E}|\times D + L \times  |\mathcal{V}| \times D^2}_{\text{vanilla GCN}} + \underbrace{k\times|\mathcal{E}| \times D \times D^s}_{\text{sparsifier experts}} + \underbrace{k \left(\frac{|\mathcal{E}|}{|\mathcal{V}|}\right)^3 + k \left(\frac{|\mathcal{E}|}{|\mathcal{V}|}\right)^2  p}_{\text{graph mixture}}\right).
\end{equation}

To empirically verify that \ourmethod does not impose excessive computational burdens on GNN backbones, we conduct experiments in \Cref{sec:exp_eff} to compare the per-epoch time efficiency metric of \ourmethod with other sparsifiers.

\section{Experimental Details}\label{app:exp_details}

\subsection{Dataset Statistics}\label{app:dataset}

We conclude the dataset statistics in Tab.~\ref{tab:dataset}

\begin{table}[htb]
\centering
\caption{Graph datasets statistics.}
\label{tab:dataset}
{\small
\begin{tabular}{c  |c |c |c  |c | c  }
\toprule
Dataset  & $\#$Graph &  $\#$Node & $\#$Edge & $\#$Classes & Metric  \\ 
\midrule
\textsc{Ogbn-ArXiv} & 1 & 169,343 & 1,166,243   & 40 & Accuracy \\
\textsc{Ogbn-Proteins} & 1 & 132,534 & 39,561,252   & 2 & ROC-AUC \\
\textsc{Ogbn-Products} & 1 &{2,449,029} & {61,859,140}   & {47} & {Accuracy} \\
\midrule
\textsc{Ogbg-PPA} & 158,100 &243.4	 & 2,266.1   & {47} & {Accuracy} \\
\midrule
\textsc{mnist} & 70,100 & 50.5	 &  564.5& {10} & {Accuracy} \\
\textsc{cifar-10} &60,000 & 117.6	 & 914.0 & {10} & {Accuracy} \\
\bottomrule
\end{tabular}}
\vspace{-2mm}
\end{table}

\subsection{Evaluation Metrics}

 Accuracy represents the ratio of correctly predicted outcomes to the total predictions made. The ROC-AUC (Receiver Operating Characteristic-Area Under the Curve) value quantifies the probability that a randomly selected positive example will have a higher rank than a randomly selected negative example. 

\subsection{Dataset Splits} 

For \textbf{node-level tasks}, the data splits for \textsc{Ogbn-ArXiv}, \textsc{Ogbn-Proteins}, and \textsc{Ogbn-Products} were provided by the benchmark~\citep{hu2020open}. Specifically, for \textsc{Ogbn-ArXiv}, we train on papers published until 2017, validate on papers from 2018 and test on those published since 2019. For \textsc{Ogbn-Proteins}, protein nodes were segregated into training, validation, and test sets based on their species of origin. {For \textsc{Ogbn-Products}, we sort the products according to their sales ranking and use the top 8\% for training, next top 2\% for validation, and the rest for testing}.

For \textbf{graph-level tasks}, we follow \citep{hu2020open} for \textsc{Ogbg-PPA}. Concretely, we adopt the species split, where the neighborhood graphs in the validation and test sets are extracted from protein association networks of species not encountered during training but belonging to one of the 37 taxonomic groups. This split stress-tests the model's capacity to extract graph features crucial for predicting taxonomic groups, enhancing biological understanding of protein associations. For \textsc{mnist} and \textsc{cifar-10}, consistent with \citep{dwivedi2020benchmarking}, we split them to 55000 train/5000 validation/10000 test for MNIST, and 45000 train/5000 validation/10000 test for CIFAR10, respectively.
We report the test accuracy at the epoch with the best validation accuracy.


\subsection{Parameter Setting}\label{app:exp_param}

\paragraph{Backbone Parameters} For node classification backbones, we utilize a 3-layer GraphSAGE with $\texttt{hidden\_dim}\in\{128,256\}$. As for DeeperGCN, we set $\texttt{layer\_num}=28$, $\texttt{block}=\texttt{res+}$, $\texttt{hidden\_dim}=64$. The other configurations are the same as in \url{https://github.com/lightaime/deep_gcns_torch/tree/master/examples/ogb/ogbn_proteins}. For graph classification backbones, we leverage a 4-layer PNA with $\texttt{hidden\_dim}=300$. Rest configurations are the same as in \url{https://github.com/lukecavabarrett/pna}.

\paragraph{MoG parameters} We adopt the $m=4$ sparsity criteria outlined in \Cref{sec:sparsifier}, assigning $n=3$ different sparsity levels $\{s_1, s_2, s_3\}$ to each criterion, resulting in a total of $K=m \times n=12$ experts. We select $k=2$ sparsifier experts for each node, and set the loss scaling factor $\lambda=1e-2$ across all datasets and backbones. 

All the experiments are conducted on NVIDIA Tesla V100 (32GB GPU), using PyTorch and PyTorch Geometric framework.

\subsection{Sparsifier Baseline Configurations}\label{app:exp_sparsifier}

\begin{itemize}[leftmargin=*]
    \item Topology-based sparsification
    \begin{itemize}[leftmargin=*]
        \item \textbf{Rank Degree}~\citep{talati2022}: The Rank Degree sparsifier initiates by selecting a random set of "seed" vertices. Then, the vertices with connections to these seed vertices are ranked based on their degree in descending order. Subsequently, the edges linking each seed vertex to its top-ranked neighbors are chosen and integrated into the sparsified graph. The newly added nodes in the graph act as new seeds for identifying additional edges. This iterative process continues until the target sparsification limit is attained. We utilize the implementation in~\citep{chen2023demystifying}.
        \item \textbf{Local Degree}~\citep{hamann2016structure}: Local Degree sparsifier, similar to Rank Degree, incorporates edges to the top $\operatorname{deg}(v)^\alpha$ neighbors ranked by their degree in descending order, where $\alpha \in [0, 1]$ represents the degree of sparsification.
        \item \textbf{Forest Fire}~\citep{Leskovec2006}: Forest fire assembles “burning” through edges probabilistically, and we use the implementation in \citep{staudt2016networkit}.
        \item \textbf{G-Spar}~\citep{murphy1996finley}: G-Spar sorts the Jaccard scores globally and then selects the edges with the highest similarity score. We opt for the code from \citep{staudt2016networkit}.
        \item  \textbf{Local Similarity}~\citep{satuluri2011local}: Local Similarity ranks edges using the Jaccard score and computes $\log(\operatorname{rank}(e_{ij}))/\log(\operatorname{deg}(e_{ij}))$ as the similarity score, and selects edges with the highest similarity scores. We utilize the implementation in~\citep{chen2023demystifying}.
        \item \textbf{SCAN}~\citep{spielman2008graph}: SCAN uses structural similarity (called SCAN similarity) measures to detect clusters, hubs, and outliers. We utilize the implementation in~\citep{chen2023demystifying}    
        \item \textbf{DSpar}~\citep{liu2023dspar}: DSpar is an extension of effective resistance sparsifier, which aims to reduce the high computational budget of calculating effective resistance through an unbiased approximation. We adopt their official implementation~\citep{liu2023dspar}.
    \end{itemize}
    \item Semantic-based sparsification
    \begin{itemize}
        \item \textbf{UGS}~\citep{chen2021unified}: We utilize the official implementation from the authors. Notably, UGS was originally designed for joint pruning of model parameters and edges. Specifically, it sets separate pruning parameters for parameters and edges, namely the weight pruning ratio $p_\theta$ and the graph pruning ratio $p_g$. In each iteration, a corresponding proportion of parameters/edges is pruned. For a fairer comparison, we set $p_\theta=0\%$, while maintaining $p_g\in\{5\%,10\}$ (consistent with the original paper).
        \item \textbf{GEBT}~\citep{you2022early}: GEBT, for the first time, discovered the existence of graph early-bird (GEB) tickets that emerge at the very early stage when sparsifying GCN graphs. \citep{you2022early} has proposed two variants of graph early bird tickets, and we opt for the graph-sparsification-only version, dubbed GEB Ticket Identification.
        \item \textbf{Meta-gradient sparsifier}~\citep{wan2021edge}: The Meta-gradient sparsifier prunes edges based on their meta-gradient importance scores, assessed over multiple training epochs. Since no official implementation is provided, we carefully replicated the results following the guidelines in the original paper.
        \item \textbf{ACE-GLT}~\citep{wang2023adversarial}: ACE-GLT inherits the iterative magnitude pruning (IMP) paradigm from UGS. Going beyond UGS, it suggested mining valuable information from pruned edges/weights after each round of IMP, which in the meanwhile doubled the computational cost of IMP. We utilize the official implementation provided by \citep{wang2023adversarial}, and set $p_\theta=0\%, p_g\in\{5\%,10\}$.
        \item \textbf{WD-GLT}~\citep{hui2022rethinking}: WD-GLT also inherits the iterative magnitude pruning paradigm from UGS, so we also set $p_\theta=0\%, p_g\in\{5\%,10\%\}$ across all datasets and backbones. The perturbation ratio $\alpha$ is tuned among $\{0,1\}$. Since no official implementation is provided, we carefully reproduced the results according to the original paper.
        \item \textbf{AdaGLT}~\citep{zhang2024graph}: AdaGLT revolutionizes the original IMP-based graph lottery ticket methodology into an adaptive, dynamic, and automated approach, proficient in identifying sparse graphs with layer-adaptive structures. We fix $\eta_\theta=0\%, \eta_g\in\{1e-6,1e-5,1e-4,1e-3,1e-2\},\omega=2$ across all datasets and backbones.

    \end{itemize}
\end{itemize}

\subsection{Adjusting Graph Sparsity}\label{app:adjust}

In \Cref{tab:recipe}, we provide detailed guidelines on how to achieve the desired global sparsity by adjusting the three sparsity levels $\{s_1, s_2, s_3\}$ in \ourmethod across six datasets.

\begin{table}[!htbp]
    \centering
    \caption{The recipe for adjusting graph sparsity via different sparsifier combinations. }
    \label{tab:recipe}
    \setlength{\tabcolsep}{10pt}
    \begin{tabular}{c|ccc|cc}
        \toprule
        Datasets & $1-s_1$ & $1-s_2$ & $1-s_3$  & $k$ & $1-s\%$ \\
        \midrule
        \multirow{4}{*}{\textsc{ogbn-arxiv}} & 1 &0.9 &0.8 & 2 & [88.0\%,90.9\%] \\
        & 0.8 &0.7& 0.5 & 2 & [69.0\%,73.2\%] \\
         & 0.6& 0.5& 0.3 & 2 & [49.5\%,52.7\%] \\
         & 0.5 &0.3 &0.15 & 2 & [27.1\%, 31.6\%] \\
        \multirow{4}{*}{\textsc{ogbn-proteins}} & 1 &0.9& 0.8 & 2 & [86.1\%,89.3\%] \\
         & 0.8 &0.7& 0.6 & 2 & [65.1\%,69.2\%] \\
        & 0.6 &0.5 &0.4 & 2 &[45.2\%,49.3\%]\\
         & 0.4 &0.3& 0.2 & 2 & [29.2\%,31.1\%] \\        
         \multirow{4}{*}{\textsc{ogbn-products}} & 1 &0.9& 0.8 & 2 & [90.1\%,93.2\%] \\
         & 0.8 &0.7& 0.6 & 2 & [69.3\%,72.0\%] \\
        & 0.6 &0.5 &0.4 & 2 &[51.5\%,54.9\%]\\
         & 0.4 &0.3& 0.2 & 2 & [28.7\%,36.0\%] \\
        \multirow{4}{*}{\textsc{mnist}} & 1 & 0.85 &0.8 & 2 & [90.4\%,92.7\%] \\
         & 0.8 &0.5 &0.4 & 2 & [67.1\%,68.3\%] \\
        & 0.6 &0.3 &0.2 & 2 & [46.2\%,49.3\%] \\
        & 0.35 &0.1 & 0.1 & 2 & [29.8\%,31.3\%]\\
        \multirow{4}{*}{\textsc{cifar-10}} & 1 & 0.85 &0.8 & 2 & [90.6\%,93.7\%] \\
         & 0.8 &0.5 &0.4 & 2 & [67.5\%,69.9\%] \\
        & 0.6 &0.3 &0.2 & 2 & [47.7\%,49.3\%] \\
        & 0.35 &0.1 & 0.1 & 2 & [30.1\%,31.3\%] \\
        \multirow{4}{*}{\textsc{ogbg-ppa}} & 0.95& 0.9& 0.8 & 2 & [86.5\%,88.9\%] \\
        & 0.8 &0.65& 0.6 & 2 &[68.0\%,70.1\%]\\
        & 0.6& 0.5& 0.3 & 2 & [47.8\%,48.9\%] \\
         & 0.4& 0.3& 0.15 & 2 & [30.1\%,33.6\%] \\
        \bottomrule
    \end{tabular}
    
\end{table}

\section{Addtional Experiment Results}\label{app:exp_add}

\subsection{Results for RQ1}

We report the performances of \ourmethod and other sparsifiers on \textsc{ogbn-products} in \Cref{tab:rq1_products}.

\begin{table*}[!htbp]
    \caption{Node classification performance comparison to state-of-the-art sparsification methods. All methods are trained using \textbf{GraphSAGE}, and the reported metrics represent the average of \textbf{five runs}. We denote methods with $\dag$ that do not have precise control over sparsity; their performance is reported around the target sparsity $\pm 2\%$. We do not report results for sparsifiers like ER for OOT issues and those like UGS for their infeasibility in inductive settings (mini-batch training). 
    }
\label{tab:rq1_products}
    \centering
    \footnotesize
    \setlength{\tabcolsep}{3pt}
    \begin{tabular}{cc|cccc}
    \toprule
    \multirow{3}{*}{} & Dataset  & \multicolumn{4}{c}{\textsc{ogbn-products} (Accuracy $\uparrow$)}  \\
    \midrule
    & Sparsity \% & 10  & 30 & 50 & 70  \\ \midrule
    \parbox[t]{4mm}{\multirow{8}{*}{\rotatebox[origin=c]{90}{Topology}}}
    & Random
    & 76.99\blue{1.05} & 74.21\blue{3.83} & 71.08\blue{6.96} & 67.24\blue{10.80}\\
    
    & Rank Degree$^{\dag}$~\citep{voudigari2016rank}  
    & 76.08\blue{1.96} & 74.26\blue{3.89} & 71.85\blue{6.19} & 70.66\blue{7.38} \\
    
    & Local Degree$^{\dag}$~\citep{hamann2016structure}   
    & 77.19\blue{1.58} & 76.40\blue{1.64} &  72.77\blue{5.27} & 72.48\blue{5.56} \\
    
    & G-Spar~\citep{murphy1996finley} 
    & 76.15\blue{1.89} & 74.20\blue{3.84} & 71.55\blue{6.49} & 69.42\blue{8.62} \\

    & LSim$^{\dag}$~\citep{satuluri2011local}
    & 77.96\blue{0.08}  & 74.98\blue{2.06} & 72.67\blue{5.37} & 70.43\blue{7.61} \\
    
    & SCAN~\citep{xu2007scan} 
    & 76.30\blue{1.74} & 74.33\blue{3.71} & 71.25\blue{6.79} & 71.12\blue{6.92}\\
    
    & DSpar~\citep{liu2023dspar}
    & 78.25\red{0.21} & 75.11\blue{2.93} & 74.57\blue{3.47} & 73.16\blue{4.88}\\
    \midrule
    \parbox[t]{4mm}{\multirow{2}{*}{\rotatebox[origin=c]{90}{Sema}}}

    & AdaGLT~\citep{zhang2024graph}
    & 78.19\red{0.15} & 77.30\blue{0.74} &  {{74.38\blue{3.66}}} & 73.04\blue{5.00}  \\
    
    & \textbf{\ourmethod (Ours)}$^{\dag}$ & \colorbox[HTML]{DAE8FC}{\textbf{78.77\red{\textbf{0.73}}}} 
    & \colorbox[HTML]{DAE8FC}{\textbf{78.15}\red{\textbf{0.11}}} 
    & \colorbox[HTML]{DAE8FC}{{\textbf{76.98}\blue{{1.06}}}} & \colorbox[HTML]{DAE8FC}{\textbf{74.91\blue{\textbf{3.17}}}}
    
      \\

    \midrule
    \multicolumn{2}{c|}{Whole Dataset} & \multicolumn{4}{c}{78.04$_{\pm0.31}$}  \\
    \bottomrule
    \end{tabular}

\end{table*}

\begin{table*}[!htbp]
    \caption{Node classification performance comparison to state-of-the-art sparsification methods. All methods are trained using \textbf{DeeperGCN}, and the reported metrics represent the average of \textbf{five runs}. We denote methods with $\dag$ that do not have precise control over sparsity; their performance is reported around the target sparsity $\pm 2\%$. ``OOM'' and ``OOT'' denotes out-of-memory and out-of-time, respectively. 
    }
\label{tab:rq1_arxiv_2}
    \centering
    \footnotesize
    \setlength{\tabcolsep}{1pt}
\begin{tabular}{cc|cccc}
    \toprule
    \multirow{3}{*}{} & Dataset  & \multicolumn{4}{c}{\textsc{ogbn-arxiv} (Accuracy $\uparrow$)}  \\
    \midrule
    & Sparsity \% & 10  & 30 & 50 & 70  \\ \midrule
    \parbox[t]{4mm}{\multirow{9}{*}{\rotatebox[origin=c]{90}{Topology-guided}}}
    & Random
    & 70.66\blue{1.28} & 68.74\blue{3.20} & 65.38\blue{6.56} & 63.55\blue{8.39}\\
    
    & Rank Degree$^{\dag}$~\citep{voudigari2016rank}  
    & 69.44\blue{2.50} & 67.82\blue{4.12} & 65.08\blue{6.86} & 63.19\blue{8.75} \\
    
    & Local Degree$^{\dag}$~\citep{hamann2016structure}   
    & 68.77\blue{3.17} & 67.92\blue{4.02} & 66.10\blue{5.84} & 65.97\blue{5.97} \\
    
    & Forest Fire$^{\dag}$~\citep{Leskovec2006}   
    & 68.70\blue{3.24}  & 68.95\blue{3.99} & 67.23\blue{4.71} & 67.29\blue{4.65}  \\
    
    & G-Spar~\citep{murphy1996finley} 
    & 70.57\blue{1.37} & 70.15\blue{1.79} & 68.77\blue{3.17} & 65.26\blue{6.68} \\

    & LSim$^{\dag}$~\citep{satuluri2011local}
    & 69.33\blue{2.61} & 67.19\blue{4.75} & 63.55\blue{8.39} & 62.20\blue{9.74} \\
    
    & SCAN~\citep{xu2007scan} 
    & 71.33\blue{0.61} & 69.22\blue{2.72} & 67.88\blue{4.06} & 64.32\blue{7.62} \\
    
    & ER~\citep{spielman2008graph}& 71.33\blue{0.61} & 69.65\blue{2.29} & 69.08\blue{2.86} & 67.10\blue{4.84} \\
    
    & DSpar~\citep{liu2023dspar}
    & 71.65\blue{0.29} & 70.66\blue{1.28} & 68.03\blue{3.91} &  67.25\blue{4.69}\\
    \midrule
    \parbox[t]{4mm}{\multirow{7}{*}{\rotatebox[origin=c]{90}{Semantic-guided}}}
    & UGS$^{\dag}$~\citep{chen2021unified}
    &  72.01\red{0.93}& 70.29\blue{1.65} & 68.43\blue{3.51} & 67.85\blue{4.09} \\

    & GEBT~\citep{you2022early}
   & 70.22\blue{1.72} & 69.40\blue{2.54} & 67.84\blue{4.10} & 67.49\blue{4.45}\\

    & MGSpar~\citep{wan2021edge}
    & 70.02\blue{1.92} & 69.34\blue{2.60} & 68.02\blue{3.92} & 65.78\blue{6.16} \\

    & ACE-GLT$^{\dag}$~\citep{wang2023adversarial}
    &72.13\red{0.19}  & 71.96\red{0.02} & 69.13\blue{2.81} & 67.93\blue{4.01} \\
    
    & WD-GLT$^{\dag}$~\citep{hui2022rethinking}
    & 71.92\blue{0.02} & 70.21\blue{1.73} & 68.30\blue{3.64} & 66.57\blue{5.37} \\
    
    & AdaGLT~\citep{zhang2024graph}
    & 71.98\red{0.04} & 70.44\blue{1.50} & 69.15\blue{2.79} & 68.05\blue{3.89} \\
    
    & \textbf{\ourmethod (Ours)}$^{\dag}$ & \colorbox[HTML]{DAE8FC}{\textbf{72.08\red{\textbf{0.14}}}} 
    & \colorbox[HTML]{DAE8FC}{\textbf{71.98}\red{\textbf{0.05}}} 
    & \colorbox[HTML]{DAE8FC}{{\textbf{69.86}\blue{{-2.08}}}} & \colorbox[HTML]{DAE8FC}{\textbf{68.20\blue{\textbf{-3.74}}}}
   
      \\

    \midrule
    \multicolumn{2}{c|}{Whole Dataset} & \multicolumn{4}{c}{71.93$_{\pm0.04}$} \\
    \bottomrule
\end{tabular}


\end{table*}

\begin{table*}[!htbp]
    \caption{Node classification performance comparison to state-of-the-art sparsification methods. All methods are trained using \textbf{DeeperGCN}, and the reported metrics represent the average of \textbf{five runs}. We denote methods with $\dag$ that do not have precise control over sparsity; their performance is reported around the target sparsity $\pm 2\%$. ``OOM'' and ``OOT'' denotes out-of-memory and out-of-time, respectively. 
    }
\label{tab:rq1_proteins}
    \centering
    \footnotesize
    \setlength{\tabcolsep}{1pt}
\begin{tabular}{cc|cccc}
    \toprule
    \multirow{3}{*}{} & Dataset   & \multicolumn{4}{c}{\textsc{ogbn-proteins} (ROC-AUC $\uparrow$)}  \\
    \midrule
    & Sparsity \% & 10  & 30 & 50 & 70\\ \midrule
    \parbox[t]{4mm}{\multirow{9}{*}{\rotatebox[origin=c]{90}{Topology-guided}}}
    & Random
    & 80.18\blue{2.55} & 78.92\blue{3.83} & 76.57\blue{6.16} & 72.69\blue{10.04}\\
    
    & Rank Degree$^{\dag}$~\citep{voudigari2016rank}  & 80.14\blue{2.59} & 79.05\blue{3.73} & 78.59\blue{4.13} & 76.22\blue{6.51}\\
    
    & Local Degree$^{\dag}$~\citep{hamann2016structure}   
    & 79.40\blue{3.33} & 79.83\blue{3.90} & 78.50\blue{4.23} & 78.25\blue{4.48}\\
    
    & Forest Fire$^{\dag}$~\citep{Leskovec2006}   
    & 81.49\blue{1.24} & 78.47\blue{4.26} & 76.14\blue{6.59} & 73.89\blue{9.84}\\
    
    & G-Spar~\citep{murphy1996finley} 
    & 81.56\blue{1.17} & 81.12\blue{1.61} & 79.13\blue{3.60} & 77.45\blue{5.28}\\

    & LSim$^{\dag}$~\citep{satuluri2011local}
    & 80.30\blue{2.43} & 79.19\blue{3.54} & 77.13\blue{5.60} & 77.85\blue{4.88}\\
    
    & SCAN~\citep{xu2007scan} 
    & 81.60\blue{1.13} & 80.19\blue{2.54} & 81.53\blue{1.20} & 78.58\blue{4.15}\\
    
    & ER~\citep{spielman2008graph}& \multicolumn{4}{c}{OOT} \\
    
    & DSpar~\citep{liu2023dspar}
     & 81.46\blue{1.27} & 80.57\blue{2.16} & 77.41\blue{5.32} & 75.35\blue{7.39}\\
     
    \midrule
    \parbox[t]{4mm}{\multirow{7}{*}{\rotatebox[origin=c]{90}{Semantic-guided}}}
    & UGS$^{\dag}$~\citep{chen2021unified}
    & 82.33\blue{0.40} & 81.54\blue{1.19} & 78.75\blue{4.98} & 76.40\blue{6.33}\\

    & GEBT~\citep{you2022early}
     & 80.74\blue{2.99} & 80.22\blue{2.51} & 79.81\blue{3.92} & 76.05\blue{6.68} \\

    & MGSpar~\citep{wan2021edge}
    & \multicolumn{4}{c}{OOM} \\

    & ACE-GLT$^{\dag}$~\citep{wang2023adversarial}
     & 82.93\red{0.80} & 82.01\blue{0.72} & 81.05\blue{1.68} & 75.92\blue{6.81} \\
    
    & WD-GLT$^{\dag}$~\citep{hui2022rethinking}
    & \multicolumn{4}{c}{OOM}\\
    
    & AdaGLT~\citep{zhang2024graph}
     & 82.60\blue{0.13} & 82.76\red{0.97} &  {{80.55\blue{2.18}}} & 78.42\blue{4.31}  \\
    
    & \textbf{\ourmethod (Ours)}$^{\dag}$ 
    & \colorbox[HTML]{DAE8FC}{\textbf{83.32\red{\textbf{0.41}}}} 
    & \colorbox[HTML]{DAE8FC}{\textbf{82.14}\blue{\textbf{0.59}}} 
    &  \colorbox[HTML]{DAE8FC}{\textbf{81.92\blue{\textbf{0.81}}}} & \colorbox[HTML]{DAE8FC}{\textbf{80.90\blue{\textbf{1.83}}}}
      \\

    \midrule
    \multicolumn{2}{c|}{Whole Dataset} & \multicolumn{4}{c}{82.73$_{\pm0.02}$} \\
    \bottomrule
    \end{tabular}


\end{table*}

\begin{table*}[!htbp]
    \caption{Graph classification performance comparison to state-of-the-art sparsification methods. All methods are trained using \textbf{PNA}, and the reported metrics represent the average of \textbf{five runs}. We denote methods with $\dag$ that do not have precise control over sparsity; their performance is reported around the target sparsity $\pm 2\%$.  
    }
\label{tab:rq1_cifar}
    \centering
    \footnotesize
    \setlength{\tabcolsep}{3pt}
\begin{tabular}{cc|cccc}
    \toprule
    \multirow{3}{*}{} & Dataset  & \multicolumn{4}{c}{\textsc{cifar-10} (Accuracy $\uparrow$)}  \\
    \midrule
    & Sparsity \% & 10  & 30 & 50 & 70  \\ \midrule
    \parbox[t]{4mm}{\multirow{8}{*}{\rotatebox[origin=c]{90}{Topology}}}
    & Random
    & 68.04\blue{1.70} & 66.81\blue{2.93} & 65.35\blue{4.39} & 62.14\blue{7.60}\\
    
    & Rank Degree$^{\dag}$~\citep{voudigari2016rank}  
    & 68.27\blue{1.77} & 67.14\blue{2.60} & 64.05\blue{5.69} & 60.22\blue{9.52} \\
    
    & Local Degree$^{\dag}$~\citep{hamann2016structure}  
    & 68.10\blue{1.64} & 67.29\blue{2.45} &  64.96\blue{4.78} & 61.77\blue{8.97} \\
    
    & G-Spar~\citep{murphy1996finley} 
    & 67.13\blue{2.61} & 65.06\blue{4.68} & 64.86\blue{4.88} & 62.92\blue{6.82} \\

    & LSim$^{\dag}$~\citep{satuluri2011local}
    & 69.75\red{0.01}  & 67.33\blue{2.41} & 66.58\blue{3.16} & 64.86\blue{4.88} \\
    
    & SCAN~\citep{xu2007scan} 
    & 68.25\blue{1.49} & 66.11\blue{3.63} & 64.59\blue{5.15} & 63.20\blue{6.54}\\
    
    & DSpar~\citep{liu2023dspar}
    & 68.94\red{0.53} & 66.80\blue{2.94} & 64.87\blue{4.87} & 64.10\blue{5.64}\\
    \midrule
    \parbox[t]{4mm}{\multirow{2}{*}{\rotatebox[origin=c]{90}{Sema}}}

    & AdaGLT~\citep{zhang2024graph}
    & 69.77\red{0.02} &  67.97\blue{1.78} &  {{65.06\blue{4.68}}} & 64.22\blue{5.52}  \\
    
    & \textbf{\ourmethod (Ours)}$^{\dag}$ & \colorbox[HTML]{DAE8FC}{\textbf{70.04\red{\textbf{0.30}}}} 
    & \colorbox[HTML]{DAE8FC}{\textbf{69.80}\red{\textbf{-0.94}}} 
    & \colorbox[HTML]{DAE8FC}{{\textbf{68.28}\blue{-1.46}}} & \colorbox[HTML]{DAE8FC}{\textbf{66.55\blue{-3.19}}}
    
      \\

    \midrule
    \multicolumn{2}{c|}{Whole Dataset} & \multicolumn{4}{c}{69.74$_{\pm0.17}$}  \\
    \bottomrule
\end{tabular}


\end{table*}

\subsection{Sensitivity analysis of parameter k}
\label{app:sen_proteins_rev}
Based on the experiments in \Cref{sec:sensitivity}, we further provide sensitivity analysis results on \textsc{ogbn-proteins}+RevGNN, as shown in \Cref{tab:sen_proteins_rev}. It can be observed that MoG achieves peak performance at $k \in \{2,3\}$ and begins to decline after $k \geq 4$, which is consistent with our finding in Observation 6.

\begin{table*}[!htbp]
\centering
\vspace{-1em}
\caption{Sensitivity analysis of parameter $k$ when applying MoG to \textsc{ogbn-proteins}+RevGNN.}
\begin{tabular}{c|ccccc}
\toprule
$k$   & 1    & 2    & 3    & 4    & 5    \\
\midrule
MoG   & 88.37 & 89.04 & 89.09 & 88.55 & 88.20 \\
\bottomrule
\end{tabular}
\label{tab:sen_proteins_rev}
\end{table*}

We also reported the impact of selecting different values of $k$ on the per-epoch training time and inference time, when applyng MoG to \textsc{ogbn-arxiv}+GraphSAGE in \Cref{tab:sen_arxiv_sage}. It can be observed that although the training and inference cost of MoG increases as the number of selected experts increases, this additional cost is not significant: when $k$ doubles from 2 to 4, the inference time only increases by 23\%. More importantly, we can already achieve optimal performance with $k \in \{2,3\}$, so there is no need to select too many experts, therefore avoiding significant inference delay.

\begin{table*}[!htbp]
\centering
\caption{Sensitivity analysis of parameter $k$ when applying MoG to \textsc{ogbn-arxiv}+GraphSAGE.}
\begin{tabular}{c|c|c|c|c}
\toprule
Sparsity & Selected Expert $k$ & Per-Epoch Time & Inference Time & Acc. \\
\midrule
30\%     & 2                    & 0.213          & 0.140          & 70.53 \\
30\%     & 3                    & 0.241          & 0.161          & 70.48 \\
30\%     & 4                    & 0.266          & 0.173          & 70.13 \\
30\%     & 5                    & 0.279          & 0.190          & 69.57 \\
\bottomrule
\end{tabular}
\label{tab:sen_arxiv_sage}
\end{table*}

\subsection{Ablation study on the noise control of the router networ}
\label{app:ablation}
\begin{table}[!ht]\scriptsize
\centering
\vspace{-0.7em}
\caption{Ablation study on the noise control of the router network $\Psi$. $\epsilon \sim \mathcal{N}(0,\mathbf{I})$ corresponds to the original setting in our paper,  $\epsilon =0$ corresponds to completely remove the noise modeling, and $\epsilon =0.2$ corresponds to fixing the noise coefficient.}
\label{tab:ablation}
\setlength{\tabcolsep}{6pt}
\begin{tabular}{l|cccc}
\toprule
 Sparsity   & Train Acc & Valid Acc & Test Acc  & $k$ \\
 \midrule
 \multicolumn{5}{c}{$\epsilon \sim \mathcal{N}(0,\mathbf{I})$}\\
\midrule
10\% & 77.20 & 72.68  & 71.93 & 3  \\
30\% & 76.03 & 71.90  & 70.53 &  3 \\
50\% & 72.45 & 69.54  & 69.06 &  3 \\
\midrule
 \multicolumn{5}{c}{$\epsilon =0$}\\
 \midrule
 10\% & 76.87 & 72.05  &  71.27 & 3  \\
30\% & 75.99 &71.15 &  70.14&  3 \\
50\% & 72.09  & 68.34  & 67.05 &  3 \\
\midrule
 \multicolumn{5}{c}{$\epsilon =0.2$}\\
 \midrule
  10\% &  76.98 & 72.22  & 71.75  & 3  \\
30\% & 75.98  &  71.48 &  70.27 &  3 \\
50\% & 73.15 & 69.84  & 68.45&  3 \\

\bottomrule
\end{tabular}
\vspace{-1.6em}
\end{table}

We test three different settings of epsilon on GraphSAGE+Ogbn-Arxiv: (1) $\epsilon\sim\mathcal{N}(0,\mathbf{I})$, (2) $\epsilon=0$, and (3) $\epsilon=0.2$, and report their performance under different sparsity levels in \Cref{tab:ablation}. We can see that trainable noisy parameters always bring the greatest performance gain to the model, which is consistent with previous practices in MoE that the randomness in the gating network is beneficial.

\section{Broader Impact} MoG, as a novel concept in graph sparsification, holds vast potential for general application. It allows for the sparsification of each node based on its specific circumstances, making it well-suited to meet the demands of complex real-world scenarios such as financial fraud detection and online recommender systems, which require customized approaches. More importantly, MoG provides a selectable pool for future sparsification, enabling various pruning algorithms to collaborate and enhance the representational capabilities of graphs.

\end{document}